\pdfoutput=1

\documentclass[11pt]{article}

\usepackage[final]{acl}

\usepackage{times}
\usepackage{latexsym}

\usepackage[T1]{fontenc}

\usepackage[utf8]{inputenc}

\usepackage{hyperref}       
\usepackage{url}            
\usepackage{booktabs}       
\usepackage{amsfonts}       
\usepackage{nicefrac}       
\usepackage{microtype}      
\usepackage{color}
\usepackage{algorithm,algpseudocode}
\usepackage{caption}
\usepackage{mwe}
\usepackage{lipsum}

\usepackage{graphicx}
\usepackage{amsmath}
\usepackage{amssymb}
\usepackage{float}
\usepackage{stfloats}
\usepackage{bbm}
\usepackage{dsfont}
\usepackage{algorithm}
\usepackage{array}
\usepackage[caption=false,font=normalsize,labelfont=sf,textfont=sf]{subfig}
\usepackage{textcomp}
\usepackage{verbatim}
\usepackage{multirow}
\usepackage{graphicx}
\usepackage{pifont}
\usepackage{color}
\usepackage[hang,flushmargin]{footmisc} 
\usepackage{booktabs}
\usepackage[normalem]{ulem}

\usepackage{soul}
\usepackage{float}
\usepackage{tcolorbox}

\usepackage{colortbl}
\usepackage{xcolor}

\soulregister{\cite}7
\soulregister{\ref}7
\soulregister\eqref7

\usepackage{colortbl}
\definecolor{azure}{rgb}{0.94, 1.0, 1.0}
\definecolor{oldlace}{rgb}{0.99, 0.96, 0.9}
\definecolor{pearl}{rgb}{0.94, 0.92, 0.84}
\definecolor{seashell}{rgb}{1.0, 0.96, 0.93}
\definecolor{silver}{rgb}{0.75, 0.75, 0.75}
\definecolor{platinum}{rgb}{0.9, 0.89, 0.89}
\definecolor{almond}{rgb}{0.94, 0.87, 0.8}
\definecolor{beige}{RGB}{245, 222, 179}

\newcommand{\circledone}{\ding{192}}%
\newcommand{\circledtwo}{\ding{193}}%
\newcommand{\circledthree}{\ding{194}}%
\newcommand{\circledfour}{\ding{195}}%
\newcommand{\circledfive}{\ding{196}}%

\title{\textsc{MlingConf}: A Comprehensive Study of Multilingual Confidence Estimation on Large Language Models}

\author{
\bf Boyang Xue$^{\spadesuit}$\thanks{\ Equal contributions.}~, 
Hongru Wang$^{\spadesuit*}$, 
Rui Wang$^{\spadesuit}$,
Sheng Wang$^{\clubsuit}$,
Zezhong Wang$^{\spadesuit}$ \\
\bf Yiming Du$^{\spadesuit}$, 
Bin Liang$^{\spadesuit}$, 
Wenxuan Zhang$^{\text{\ding{117}}}$,
Kam-Fai Wong$^{{\spadesuit}{\text{\ding{118}}}}$\thanks{~Corresponding author.} \\
  $^{\spadesuit}$ The Chinese University of Hong Kong \\
  $^{\clubsuit}$ The University of Hong Kong \; $^{\text{\ding{117}}}$ Singapore University of Technology and Design \\
    $^{\text{\ding{118}}}$ MoE Key Laboratory of High Confidence Software Technologies \\
  {\tt \{byxue, hrwang, kfwong\}@se.cuhk.edu.hk}}

\begin{document}

\maketitle
\begin{abstract}

The tendency of Large Language Models (LLMs) to generate hallucinations raises concerns regarding their reliability.
Therefore, confidence estimations indicating the extent of trustworthiness of the generations become essential.
However, current LLM confidence estimations in languages other than English remain underexplored.
This paper addresses this gap by introducing a comprehensive investigation of \textbf Multi\textbf{ling}ual \textbf{Conf}idence estimation (\textsc{MlingConf}) on LLMs, focusing on both language-agnostic (LA) and language-specific (LS) tasks to explore the performance and language dominance effects of multilingual confidence estimations on different tasks.
The benchmark comprises four meticulously checked and human-evaluated high-quality multilingual datasets for LA tasks and one for the LS task tailored to specific social, cultural, and geographical contexts of a language.
Our experiments reveal that on LA tasks \textit{English} exhibits notable linguistic dominance in confidence estimations than other languages, while on LS tasks, using question-related language to prompt LLMs demonstrates better linguistic dominance in multilingual confidence estimations.
The phenomena inspire a simple yet effective native-tone prompting strategy by employing language-specific prompts for LS tasks, effectively improving LLMs' reliability and accuracy in LS scenarios.


\end{abstract}

\section{Introduction}
\label{sec:intro}

Large Language Models' (LLMs) susceptibility to generating hallucinated contents incurs concerns about unreliability in real-world applications \cite{ziwei2023survey,rawte2023survey}.
Therefore, it becomes increasingly crucial for users to directly ascertain how much they can trust a model's response.
Assessing the confidence or uncertainty of a model's output can immediately indicate the level of reliability to users, thereby playing a key role in developing trustworthy AI systems \cite{geng2023survey,kadavath2022language}.

\begin{figure}[t]
    \centering
    \includegraphics[width=0.49\textwidth]{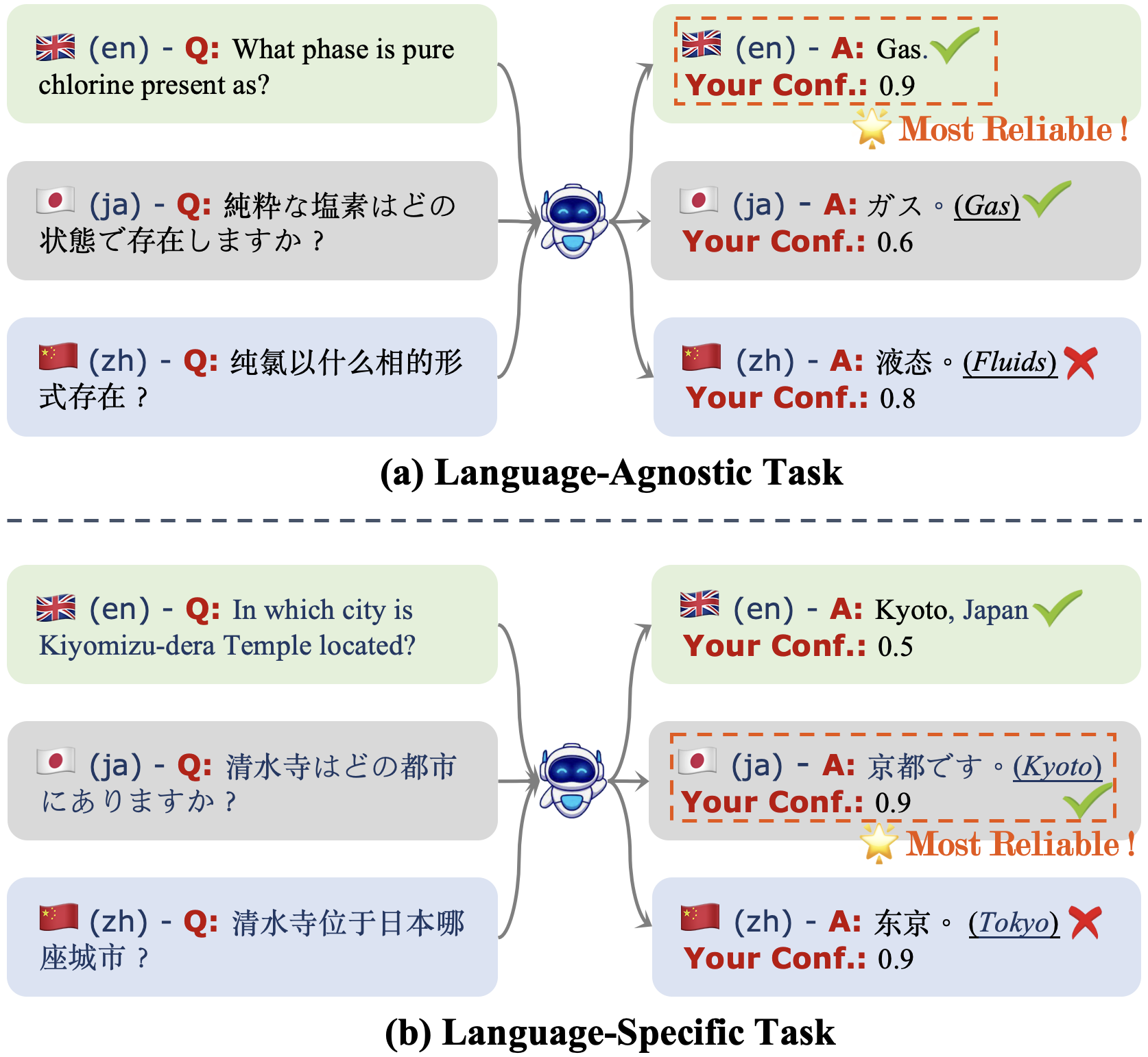}
    \caption{Examples of generations and confidence scores of {{Llama-3.1}} given the same inputs in three languages in LA and LS scenarios derived from {SciQ} and {LSQA} datasets respectively.}
    \label{fig:example}
\end{figure}


However, existing research on confidence or uncertainty estimations for LLMs has been predominantly limited to English \cite{kadavath2022language, lin2022teaching, geng2023survey, tian2023just}.
The dearth of confidence estimations in languages other than English hinders users from assessing the reliability of LLMs in non-English scenarios, restricting the LLMs' global deployment.
Due to the variations in the quantity and domain coverage of pre-training corpora across different languages, the confidence estimation ability may also presumably vary.
Therefore, the performance of confidence estimation methods primarily developed for English remains a crucial subject for explorations when applied to other languages.

Additionally, to conduct a fine-grained investigation of multilingual confidence estimations across various tasks, we divide the tasks into language-agnostic (LA) \citep{zhao2020inducing} and language-specific (LS) scenarios as in Figure \ref{fig:example} considering the effects of linguistic dominance.
Linguistic dominance refers to that one language holds a superior status over others within a specific social or cultural context \citep{blommaert2010sociolinguistics,treffers2019defines,heller2007bilingualism}, can also exist in confidence estimation ability on different languages.
In this study, the LS refers to the tasks that hold linguistic dominance caused by the knowledge domain coverage varying in different language training corpora, such as questions pertaining to social, cultural, or geographical contexts for a specific language,
while the LA involves linguistic dominance mainly caused by the quantities of training corpora, such as general knowledge, common sense, and reasoning \citep{basaj2018las,sánchez2024linguini}.

To this end, we propose a benchmark called \textsc{MlingConf} (\textbf{M}ulti\textbf{ling}ual \textbf{Conf}idence) to investigate the performance of several LLM confidence estimation methods on five languages including \textit{English}, \textit{Japanese}, \textit{Chinese}, \textit{French}, and \textit{Thai}.
First, we meticulously constructed a high-quality multilingual dataset called the MlingConf dataset for the benchmark including five datasets of different tasks in LA and LS scenarios respectively.
The LA involves four different tasks that are widely used in confidence estimation in \textit{English} \citep{kuhn2023semantic,xiong2024can} are translated into other four languages.
We also create a language-specific QA (LSQA) dataset for the LS scenario, including five subsets for the investigated five languages respectively.
Each subset comprises QA pairs about social culture, history, geography, and celebrity pertaining to the specific language.
To ensure the data quality, we conduct rigorous translation consistency checks to filter the failed samples and finally employ linguistic experts for human evaluations.

Experiments are conducted on three major LLM confidence estimation methods including probability-based \citep{vazhentsev-etal-2023-efficient,varshney2023stitch} and prompt-based confidence estimations ($p(\mathrm{True})$ \citep{kadavath2022language} and self-verbalize \citep{lin2022teaching,xiong2024can}) using the curated five multilingual datasets on several LLMs.
We evaluate the confidence estimation ability and calibration using AUROC and ECE.
Results on LA tasks suggest that prompt-based confidence estimations are preferable on LLMs with stronger instruction-following abilities, and English exhibits linguistic dominance.
Results on the LS task reveal a pronounced phenomenon of language dominance, indicating that, for questions related to specific linguistic contexts, utilizing the respective languages yields the highest accuracy and confidence estimation performance.
This observation inspires a native-tone prompting strategy: whereby, in the LS task, the relevant linguistic background of the question is first assessed, and then the corresponding language is employed to generate the response.
Compared to the use of any single language, this approach leads to significant improvements in both accuracy and confidence estimations.
Furthermore, we employ and generalize on extended confidence estimation methods and languages for both LS and LA tasks.
The results further complete and enhance the above findings and analysis to the benchmark.

The contributions are summarized as follows:

$\bullet$ To the best of our knowledge, the \textsc{MlingConf} first proposes to investigate multilingual confidence estimations with intricately constructed and expert-checked {MlingConf} datasets for both LA and LS scenarios, serving as a valuable benchmark for future works of reliable multilingual LLMs
\footnote{The codes and the {MlingConf} datasets have been released on \href{https://github.com/AmourWaltz/MlingConf}{https://github.com/AmourWaltz/MlingConf}.}.



$\bullet$ Experiments conducted on MlingConf datasets present valuable findings about confidence estimation uses in multilingual scenarios, language dominance effects of \textit{English} on LA tasks, and query-related languages on LS tasks respectively.

$\bullet$ Based on the observed linguistic dominance on LS tasks, we propose a native-tone prompting strategy, which significantly enhances the reliability and accuracy compared to the use of any single language prompts for LS tasks.

\section{{MlingConf} Dataset}
\label{sec:dataset}


Owing to the lack of multilingual resources to comprehensively exhibit confidence estimation across diverse languages, we construct a high-quality multilingual dataset called MlingConf dataset encompassing five languages: \textit{English} (\textbf{{en}}), \textit{Japanese} (\textbf{{ja}}), \textit{Chinese} (\textbf{{zh}}), \textit{French} (\textbf{{fr}}), and \textit{Thai} (\textbf{{th}}).
Specifications of the language selection in consideration of language family and resource level are demonstrated in Appendix \ref{appendix:lang}.
The MlingConf dataset includes four tasks for the language-agnostic (LA) scenario and one task for the language-specific (LS) scenario.
We specify the data source and construction process of the MlingConf dataset in Sec. \ref{ssec:source} and \ref{ssec:build} respectively.
Further dataset details and statistics can be referred to Appendix \ref{appendix:data}.


\begin{figure*}[t]
    \centering
    \includegraphics[width=0.96\textwidth]{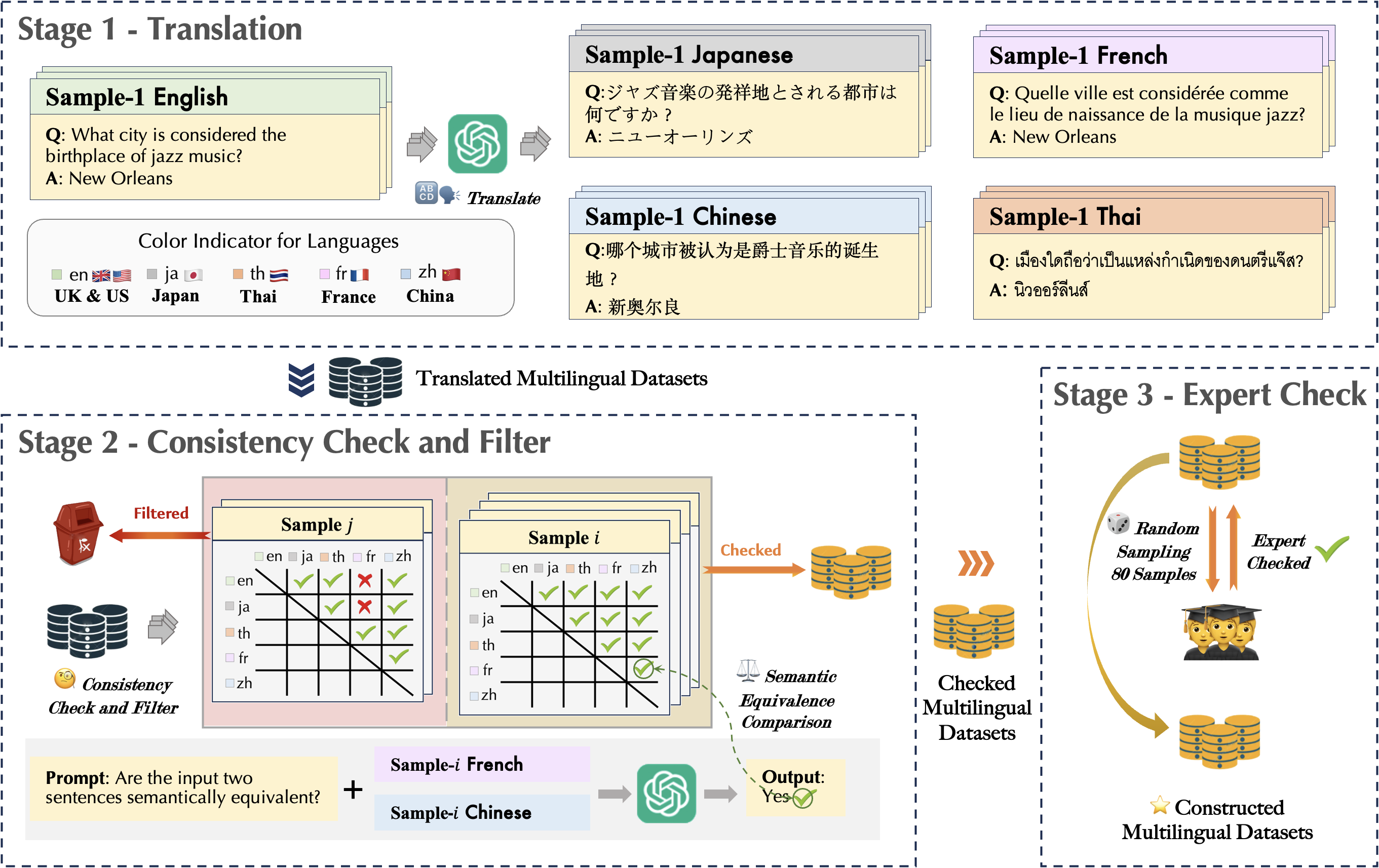}
    \caption{Three stages of MlingConf dataset construction.}
    \label{fig:data}
\end{figure*}

\subsection{Data Source}
\label{ssec:source}

\paragraph{Language-Agnostic (LA) Tasks}
For LA tasks, we employ the following four datasets that are widely used for confidence estimations in \textit{English} \citep{kuhn2023semantic,xiong2024can}.
The datasets include
1) \textbf{TriviaQA} (\textbf{TVQA}) \citep{joshi-etal-2017-triviaqa} of closed-book trivia question-answering pairs to gauge models’ factual knowledge;
2) \textbf{GSM8K} \citep{cobbe2021training} for arithmetic reasoning task of math problems;
3) \textbf{CommonsenseQA} (\textbf{CSQA}) \citep{talmor2019commonsense} of multiple-choice QA pairs requiring different types of commonsense knowledge; 
4) \textbf{SciQ} \citep{welbl2017sciq} requiring scientific professional knowledge.
All the datasets are pre-processed in standard QA format.

\paragraph{Language-Specific (LS) Tasks}
We create \textbf{Language-Specific QA} (\textbf{LSQA}) dataset pertaining to language-dominant knowledge covering specific social, geographical, and cultural language contexts for the UK \& US, France, China, Japan, and Thailand respectively.
We prompt {\textsl{GPT-4}} \cite{openai2023gpt4} \footnote{\href{https://platform.openai.com/docs/api-reference}{https://platform.openai.com/docs/api-reference}} to generate 200 questions pertaining to only one specific language contexts as a language-specific subset.
As demonstrated in Figure \ref{fig:ls_examples}, for example, all questions in \textit{Japanese} subset pertain to Japanese social culture, history, geography, celebrities, etc.


\subsection{Dataset Construction}
\label{ssec:build}

The construction of the MlingConf dataset in this study follows an elaborate three-stage procedure as delineated in Figure \ref{fig:data}.

\paragraph{Stage 1} The QA samples derived from the above datasets are translated into four languages (\textbf{{ja}}, \textbf{{th}}, \textbf{{zh}}, and \textbf{{fr}}) through {\textsl{GPT-4}} \cite{openai2023gpt4}.

\paragraph{Stage 2} We check the consistency of five translated results by comparing the semantic equivalence in pairs in $C_{5}^{2}=10$ times for each sample.
The samples with more than 2 times semantic in-equivalence are treated as noisy data and then filtered.
The changes of samples before and after consistency check and filter are in Table \ref{table:check} and more clean multilingual datasets are obtained.
Moreover, we present the number of samples for each language-specific LSQA subset as in Table \ref{table:lsqa}.

\paragraph{Stage 3} Finally, we employ several experts majoring in linguistics to examine the translation performance across 50 randomly selected samples as shown in Table \ref{table:human}.
Given the dataset obtained after Stage 2, we randomly select 50 samples from each test set and send them to language experts.
Each language expert is required to evaluate the translation results of their respective specialized language in each sample, determining whether the translation is correct (returning 1 for correct and 0 otherwise).
We calculate the translation accuracy for each language in each test set and present the results in Table \ref{table:human}. 
Human evaluation results suggest the obtained multilingual datasets are high-quality for further experiments.

\paragraph{}
For all generations of MlingConf dataset construction, the temperature $T$ is set to 0.
The translation and semantic equivalence comparison prompts are presented in Appendix \ref{appendix:prompt}.

\begin{table}[ht!]
    \centering
    \footnotesize
    \resizebox{.45\textwidth}{!}{\begin{tabular}{c|ccccc}
    \toprule
        \bf Lang. & \bf TVQA & \bf GSM8K & \bf CSQA & \bf SciQ & \bf LSQA \\
        \hline
        \hline
        \bf \texttt{zh} & 96\% & 100\% & 100\% & 98\% & 100\% \\
        \bf \texttt{ja} & 98\% & 100\% & 98\% & 96\% & 100\% \\
        \bf \texttt{fr} & 100\% & 100\% & 100\% & 98\% & 100\% \\
        \bf \texttt{th} & 96\% & 100\% & 98\% & 94\% & 100\% \\
    \bottomrule
    \end{tabular}}
    \caption{Translation accuracy evaluated by linguistic experts on 50 randomly selected samples.}
\label{table:human}
\end{table}

\begin{table}[ht!]
    \centering
    \footnotesize
    \resizebox{.47\textwidth}{!}{\begin{tabular}{c|ccccc}
    \toprule
      & \bf TVQA & \bf GSM8K & \bf CSQA & \bf SciQ & \bf LSQA \\
        \hline
        \hline
        Before & 2000 & 1319 & 1221 & 1000 & 1000 \\
        After & 1238 & 1318 & 1152 & 640 & 857 \\
    \bottomrule
    \end{tabular}}
    \caption{Number of samples before and after consistency check and filter.}
\label{table:check}
\end{table}

\begin{table}[ht!]
    \centering
    \footnotesize
    {\begin{tabular}{c|cccccc}
    \toprule
      \multirow{2}{*}{\bf LSQA} & \bf {en} & \bf {zh} & \bf {ja} & \bf {fr} & \bf {th} & \bf Total \\
        \cline{2-7}
        & 185 & 172 & 167 & 179 & 154 & 857 \\
    \bottomrule
    \end{tabular}}
    \caption{Statistics of samples for each language-specific subset of the LSQA dataset.}
\label{table:lsqa}
\end{table}

\section{Experimental Settings}
\label{sec:setting}

\subsection{Confidence Estimation Methods}

In this part, we investigate three confidence estimation methods primarily used in \textit{English} for LLMs as in Figure \ref{fig:conf}.
These methods will be also conducted in our multilingual settings.
Specifically, we denote $\mathsf{Conf}(\boldsymbol x, \boldsymbol y)$ as the confidence score associated with the output sequence $\boldsymbol y=[y_1, y_2, \dots, y_N]$ given the input context $\boldsymbol x=[x_1, x_2, \dots, x_M]$.

\paragraph{Probability-based Confidence ({Prob.}):}

The probability-based confidence is estimated by calculating the joint token-level probabilities over $\boldsymbol y$ conditioned on $\boldsymbol x$.
As longer sequences are supposed to have lower joint likelihood probabilities that shrink exponentially with length, we calculate the geometric mean by normalizing the output token probabilities which are represented as:

\vspace{-1em}
\begin{align}
\label{eq:conf_normal}
    \mathsf{Conf}(\boldsymbol x, \boldsymbol y)&=\left ( {\prod^N_ip(y_i|\boldsymbol y_{<i},\boldsymbol x)} \right )^{\frac{1}{N}}
\end{align}

\paragraph{$p(\mathrm{True})$-based Confidence ({p(True)}):}
The \textbf{$p(\mathrm{\mathrm{True}})$} confidence score is implemented by simply asking the model itself if its first proposed answer $\boldsymbol y$ to the question $\boldsymbol x$ is true \cite{kadavath2022language}, and then obtaining the probability $p(\mathrm{True})$, which can implicitly reflect self-reflected certainty.



\paragraph{Self-verbalized Confidence ({Verb.}):}
As LLMs possess good self-reflection and instruction-following abilities, 
recent works pay particular attention to linguistic confidence via prompting LLMs to express certainty in verbalized numbers or words \cite{lin2022teaching,xiong2024can}.
We adopt verbalized numerical probability in token-level space as LLM's confidence estimation.

\paragraph{}
The multilingual prompts for {{p(True)}} and {{Verb.}} are in Appendix \ref{appendix:prompt}.

\subsection{Evaluation Metrics}
\label{sec:eval}

\paragraph{Accuracy (Accu.)}
We employ a string-matching approach to evaluate the accuracy of generated answers $\boldsymbol y$ and compare them with the ground truth $\boldsymbol {\hat y}$.
Although exact matching (EM) of $\boldsymbol y\equiv \boldsymbol {\hat y}$ is widely used on GSM8K and CSQA, it always misjudges some correct answers with slight differences on closed-book QA tasks, to better assess the result accuracy (Accu.), we replace EM with a variant called positive-recall exact matching (PREM) of $\boldsymbol y\in \boldsymbol {\hat y} \vee \boldsymbol {\hat y} \in \boldsymbol y$.
Comparisons of several EM variants we tested as well as human evaluations are presented in Appendix \ref{appendix:metric}.

\paragraph{Area Under the Receiver Operator Characteristic Curve (AUROC)}
AUROC assesses the effectiveness of confidence estimation \cite{filos2019benchmarking} by quantifying how likely a randomly chosen correct answer possesses a higher confidence score than an incorrect one, yielding a score in range of [0, 1], implemented by \texttt{sklearn} toolkit \footnote{\href{https://github.com/scikit-learn/scikit-learn/blob/main/sklearn/metrics/_ranking.py}{https://github.com/scikit-learn/scikit-learn/blob/main/sklearn/metrics/\_ranking.py}}.

\paragraph{Expected Calibration Error (ECE)}
ECE gauges the calibration performance which indicates how well a model’s predicted confidence matches its actual accuracy \citep{guo2017on}.
For an expected well-calibrated AI system, samples $\boldsymbol x$ with confidence of $q$ should also have an average accuracy of $q$ on predictions $\boldsymbol y$ where $P(\boldsymbol {y}=\boldsymbol {\hat y}|\mathsf{Conf}(\boldsymbol x, \boldsymbol {y})=q)=q$ with ECE=0.
ECE is essential for reliable AI systems on prediction tasks like weather forecasting.
The smaller the ECE value, the better.
Details of the ECE calculation are presented in Appendix \ref{appendix:metric}.

\begin{table*}[t]
    \centering
    \footnotesize
    \resizebox{.95\textwidth}{!}{\begin{tabular}{c|cc|cc|cc|cc|cc|cc}
    \toprule
        \multirow{2}{*}{\bf Conf.} & \multicolumn{2}{c|}{\bf {en}} & \multicolumn{2}{c|}{\bf {zh}} & \multicolumn{2}{c|}{\bf {ja}} & \multicolumn{2}{c|}{\bf {fr}} & \multicolumn{2}{c|}{\bf {th}} & \multicolumn{2}{c}{\bf \textit{Avg.}} \\
         & \bf ARC. $\uparrow$ & \bf ECE $\downarrow$ & \bf ARC. $\uparrow$ & \bf ECE $\downarrow$ & \bf ARC. $\uparrow$ & \bf ECE $\downarrow$ & \bf ARC. $\uparrow$ & \bf ECE $\downarrow$ & \bf ARC. $\uparrow$ & \bf ECE $\downarrow$ & \bf ARC. $\uparrow$ & \bf ECE $\downarrow$ \\
        \hline
        \hline
        \rowcolor{almond}
        \multicolumn{13}{c}{\textbf{TVQA on \textsl{GPT-3.5}}} \\
        \hline
        \textbf{\textit{Prob.}} & 76.51 & 24.36 & 78.39 & 32.95 & 76.90 & \bf 28.14 & 72.14 & 27.39 & 74.30 & 40.17 & 75.65 & 30.60 \\
        \textbf{\textit{p(True)}} & 79.64 & 18.25 & \bf 82.34 & \bf 22.94 & 84.50 & 29.06 & 80.59 & \bf 20.90 & 81.22 & 40.87 & 81.66 & \bf 26.40 \\
        \textbf{\textit{Verb.}} & \bf 80.32 & \bf 16.52 & 81.76 & 24.32 & \bf 84.61 & 34.49 & \bf 83.47 & 26.53 & \bf 86.72 & \bf 38.19 & \bf 83.38 & 28.01 \\
        \hline
        \rowcolor{almond}
        \multicolumn{13}{c}{\textbf{TVQA on \textsl{Llama-3.1}}} \\
        \hline
        \textbf{\textit{Prob.}} & \bf 80.74 & \bf 10.72 & \bf 80.41 & 40.39 & \bf 88.75 & \bf 26.24 & \bf 79.05 & \bf 20.38 & \bf 89.59 & \bf 36.77 & \bf 83.71 & \bf 26.90  \\
        \textbf{\textit{p(True)}} & 68.98 & 18.35 & 68.10 & 38.19 & 52.69 & 37.85 & 62.00 & 22.04 & 60.55 & 37.01 & 62.66 & 30.69 \\
        \textbf{\textit{Verb.}} & 77.18 & 24.73 & 63.50 & \bf 37.64 & 68.91 & 34.27 & 69.90 & 25.44 & 73.22 & 40.19 & 70.54 & 32.45 \\
        \hline
        \hline
        \rowcolor{oldlace}
        \multicolumn{13}{c}{\textbf{GSM8K on \textsl{GPT-3.5}}} \\
        \hline
        \textbf{\textit{Prob.}} & 54.79 & 26.48 & 58.49 & \bf 27.19 & 57.09 & 29.46 & 57.38 & 37.29 & \bf 61.73 & \bf 41.77 & 57.90 & 32.44 \\
        \textbf{\textit{p(True)}} & \bf 65.25 & 31.88 & \bf 62.74 & 28.65 & \bf 69.75 & \bf 19.03 & 60.14 & 39.08 & 61.45 & 49.88 & \bf 63.87 & 33.70 \\
        \textbf{\textit{Verb.}} & 62.34 & \bf 22.17 & 59.25 & 28.91 & 58.34 & 26.71 & \bf 66.65 & \bf 25.14 & 54.02 & 45.63 & 60.12 & \bf 29.71 \\
        \hline
        \rowcolor{oldlace}
        \multicolumn{13}{c}{\textbf{GSM8K on \textsl{Llama-3.1}}} \\
        \hline
        \textbf{\textit{Prob.}} & \bf 65.69 & 22.33 & \bf 66.37 & 21.92 & 69.73 & 35.69 & \bf 61.07 & 29.56 & \bf 63.22 & 28.51 & \bf 65.22 & 27.60 \\
        \textbf{\textit{p(True)}} & 61.64 & \bf 14.49 & 65.83 & \bf 17.39 & \bf 71.40 & \bf 11.26 & 57.04 & \bf 8.02 & 57.31 & \bf 12.90 & 62.64 & \bf 12.81 \\
        \textbf{\textit{Verb.}} & 57.00 & 50.05 & 63.04 & 42.89 & 58.93 & 45.33 & 54.45 & 35.31 & 55.30 & 34.94 & 57.75 & 41.70 \\
        \hline
        \hline
        \rowcolor{azure}
        \multicolumn{13}{c}{\textbf{CSQA on \textsl{GPT-3.5}}} \\
        \hline
        \textbf{\textit{Prob.}} & 59.06 & 24.45 & 55.92 & 38.30 & 48.01 & 50.60 & 55.33 & 31.12 & 48.21 & 41.71 & 53.31 & 48.18 \\
        \textbf{\textit{p(True)}} & 67.13 & 19.65 & \bf 58.64 & 27.08 & 65.23 & \bf 19.24 & \bf 66.33 & 23.43 & 59.96 & 34.47 & 63.46 & 24.77 \\
        \textbf{\textit{Verb.}} & \bf 69.60 & \bf 16.84 & 54.30 & \bf 21.54 & \bf 68.34 & 19.84 & 61.87 & \bf 21.81 & \bf 68.93 & \bf 21.71 & \bf 64.73 & \bf 24.35 \\
        \hline
        \rowcolor{azure}
        \multicolumn{13}{c}{\textbf{CSQA on \textsl{Llama-3.1}}} \\
        \hline
        \textbf{\textit{Prob.}} & \bf 78.06 & \bf 13.64 & \bf 64.91 & 36.33 & \bf 75.65 & \bf 30.11 & \bf 66.12 & \bf 19.72 & \bf 77.65 & 42.18 & \bf 72.47 & \bf 28.40 \\
        \textbf{\textit{p(True)}} & 56.25 & 34.04 & 64.24 & 36.82 & 66.60 & 40.32 & 59.34 & 27.72 & 58.07 & \bf 29.91 & 60.90 & 33.76 \\
        \textbf{\textit{Verb.}} & 62.42 & 28.16 & 54.61 & \bf 28.84 & 61.06 & 37.85 & 57.97 & 23.96 & 71.91 & 37.15 & 61.39 & 31.19 \\
        \hline
        \hline
        \rowcolor{seashell}
        \multicolumn{13}{c}{\textbf{SciQ on \textsl{GPT-3.5}}} \\
        \hline
        \textbf{\textit{Prob.}} & 69.50 & 32.23 & 71.29 & 35.63 & 78.28 & 47.06 & \bf 72.66 & 34.85 & \bf 75.13 & 56.17 & 73.37 & 41.19 \\
        \textbf{\textit{p(True)}} & \bf 72.06 & 23.15 & \bf 76.18 & \bf 30.44 & \bf 80.16 & 36.18 & 71.30 & 37.85 & 68.29 & 41.25 & \bf 73.60 & 33.77 \\
        \textbf{\textit{Verb.}} & 70.18 & \bf 20.80 & 75.50 & 37.59 & 77.89 & \bf 30.33 & 69.31 & \bf 32.47 & 74.85 & \bf 41.15 & 73.55 & \bf 32.47 \\
        \hline
        \rowcolor{seashell}
        \multicolumn{13}{c}{\textbf{SciQ on \textsl{Llama-3.1}}} \\
        \hline
        \textbf{\textit{Prob.}} & \bf 74.14 & \bf 13.40 & \bf 72.09 & 32.26 & \bf 74.48 & \bf 34.21 & \bf 77.45 & 22.76 & \bf 77.61 & \bf 36.10 & \bf 75.15 & \bf 27.75 \\
        \textbf{\textit{p(True)}} & 62.38 & 19.28 & 64.89 & \bf 37.01 & 58.92 & 36.47 & 61.01 & \bf 10.72 & 51.90 & 41.06 & 59.82 & 28.91 \\
        \textbf{\textit{Verb.}} & 62.65 & 24.10 & 52.90 & 32.94 & 69.10 & 39.15 & 59.30 & 24.92 & 65.93 & 40.67 & 61.98 & 38.36 \\
        \hline
        \hline
        \rowcolor{platinum}
        \multicolumn{13}{c}{\textbf{Avg. (TVQA, GSM8K, CSQA, SciQ) on \textsl{GPT-3.5}}} \\
        \hline
        \textbf{\textit{Prob.}} & 64.96 & 26.88 & 66.02 & 33.51 & 65.07 & 38.71 & 64.37 & 32.66 & 64.88 & 44.95 & 65.06 & 35.34 \\
        \textbf{\textit{p(True)}} & 70.38 & 23.23 & \bf 69.97 & \bf 27.27 & \bf 74.91 & \bf 25.87 & 69.59 & 30.31 & 67.73 & 41.61 & \bf 70.65 & 29.66 \\
        \textbf{\textit{Verb.}} & \bf 70.61 & \bf 19.01 & 67.70 & 28.09 & 72.29 & 27.84 & \bf 70.32 & \bf 26.48 & \bf 71.13 & \bf 36.67 & 70.45 & \bf 28.64 \\
        \hline
        \textbf{\textit{Overall}} & 68.68 & 23.04 & 67.90 & 29.65 & 70.75 & 30.84 & 68.10 & 29.82 & 67.90 & 41.08 & 68.66 & 30.89 \\
        \hline
        \rowcolor{platinum}
        \multicolumn{13}{c}{\textbf{Avg. (TVQA, GSM8K, CSQA, SciQ) on \textsl{Llama-3.1}}} \\
        \hline
        \textbf{\textit{Prob.}} & \bf 74.88 & \bf 15.02 & \bf 70.94 & 32.72 & \bf 77.15 & 31.56 & \bf 70.92 & 23.10 & \bf 77.01 & 35.89 & \bf 74.14 & 27.66 \\
        \textbf{\textit{p(True)}} & 62.31 & 21.54 & 65.76 & \bf 32.21 & 62.40 & \bf 31.47 & 59.59 & \bf 17.12 & 56.95 & \bf 30.22 & 61.51 & \bf 26.54 \\
        \textbf{\textit{Verb.}} & 64.81 & 31.76 & 58.51 & 35.57 & 64.50 & 39.15 & 60.40 & 27.40 & 66.59 & 38.23 & 62.92 & 35.93 \\
        \hline
        \textbf{\textit{Overall}} & 67.26 & 22.77 & 65.10 & 33.55 & 67.97 & 34.06 & 63.72 & 22.58 & 66.85 & 34.78 & 66.19 & 29.38 \\
        \hline
    \bottomrule
    \end{tabular}}
    \caption{Experimental results of AUROC ({ARC.}) and {ECE} of three confidence estimation methods on four LA datasets on {{GPT-3.5}} and {{Llama-3.1}}.}
\label{table:la_auroc_ece}
\end{table*}

\subsection{Implementation Details}
\label{sec:implement}

Experiments are conducted on {{GPT-3.5-Turbo-0125}} ({{GPT-3.5}}) and {{Llama-3.1-8B-Instruct}} ({{Llama-3.1}}) \footnote{\href{https://huggingface.co/meta-llama/Llama-3.1-8B-Instruct}{https://huggingface.co/meta-llama/Llama-3.1-8B-Instruct}} \citep{llama3modelcard}.
We only present the results of the current most commonly used commercial {{GPT-3.5}} and open-source {{Llama-3.1}} in the main part and leave the results on some other LLMs in Appendix \ref{appendix:exp}.
Few-shot prompts containing $N_{f}$ examples are utilized for answer generation with greedy decoding which outperforms temperature decoding on knowledge tasks \citep{song2024good}. $N_f$ is set to 8 for GSM8K and 4 for others.

\section{Experiments on LA Tasks}
\label{sec:experiment}

\begin{figure*}[ht!]
    \centering
    \includegraphics[width=0.85\textwidth]{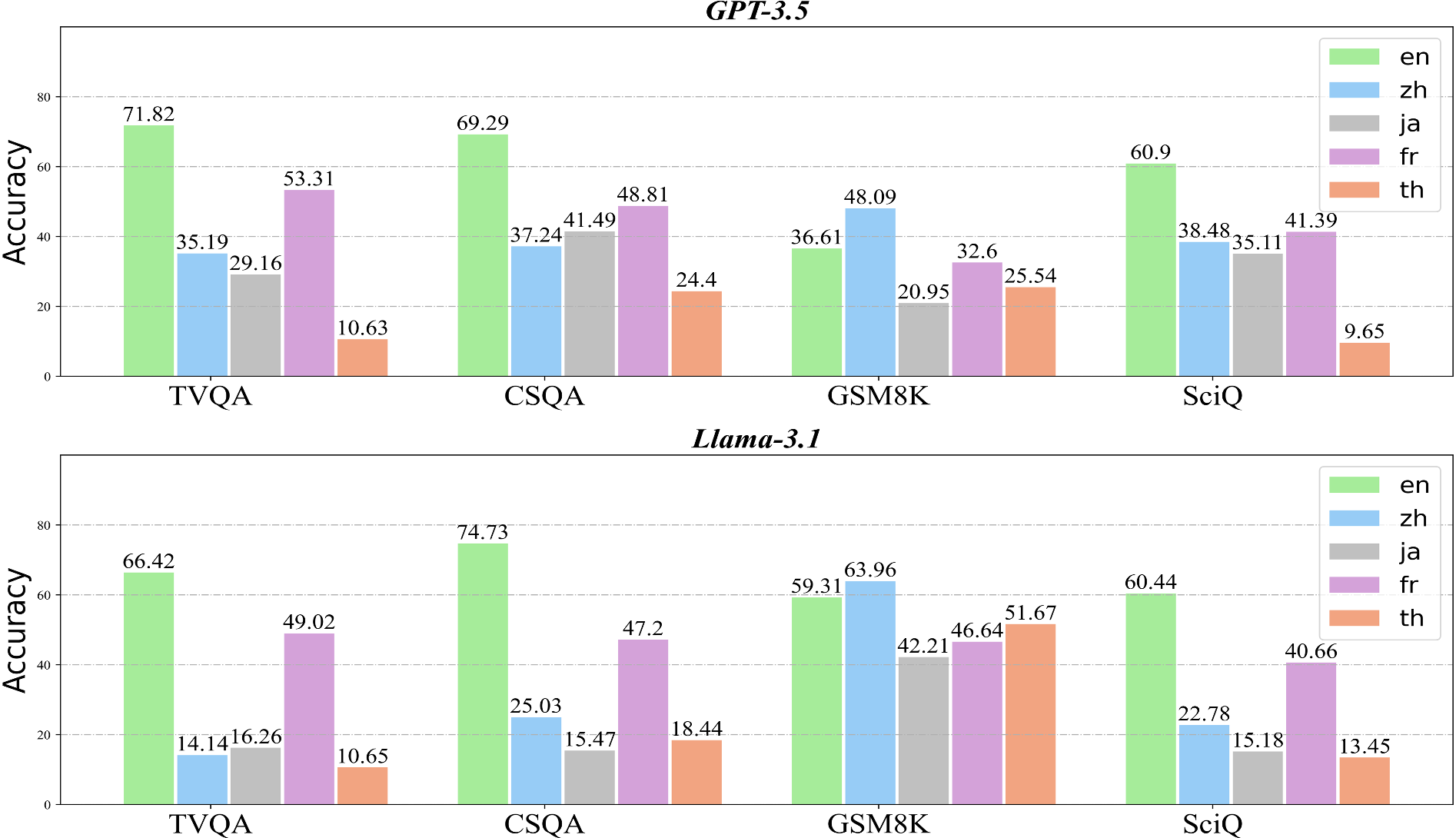}
    \caption{Experimental results of Accuracy on four LA datasets on {{GPT-3.5}} and {{Llama-3.1}}.}
    \label{fig:la_accu}
\end{figure*}

To comprehensively investigate LLMs' multilingual confidence estimations on LA tasks, as presented in Table \ref{table:la_auroc_ece} and Figure \ref{fig:la_accu},
experiments are conducted to observe performances varying in different confidence estimation methods and languages in Sec. \ref{ssec:la_res} and \ref{ssec:la_lang} respectively.






\subsection{Results regarding Different Confidence Estimations on LA Tasks}
\label{ssec:la_res}


\paragraph{Findings:}
\textbf{Applying prompt-based confidence estimations is preferable in multilingual tasks for LLM with stronger instruction-following ability.
The probability method performs better confidence estimations on relatively weak LLM.}
The findings provide a direct takeaway about selecting optimal confidence estimations of LLMs in multilingual scenarios.
In Table \ref{table:la_auroc_ece}, we highlight the supreme performance in bold among the three methods for each column of each dataset.
{On {GPT-3.5}, both {p(True)}} 
{and {Verb.}} 
{yield the superior performances than {Prob.}} 
{across all languages} averaged on four datasets
(ARC.: 70.65, 70.45 vs. 66.02; ECE: 29.66, 28.64 vs. 35.34).
{{p(True)}} and {{Verb.}} have comparable performance in ARC. scores, while {{Verb.}} is better calibrated.
In contrast, {{Prob.} shows superior performance than {p(True)}} {and {Verb.} and performs more stable on {Llama-3.1}} (ARC.: 74.14 vs. 61.51 and 62.92; ECE: 27.66 vs. 26.54 and 35.93).
{{p(True)}} demonstrates better calibration results on languages other than \textit{English}.

\paragraph{Analysis:} In Table \ref{table:la_auroc_ece}, the performance differences between two LLMs can be attributed to that {{GPT-3.5}}'s strong instruction-following abilities benefit the prompt-based multilingual confidence estimation methods {{Verb.}} and {{p(True)}}, but leading to over-confidence in output token probabilities.
In contrast, {{Llama-3.1}} cannot stably generate verbalized confidence scores and perform relatively weak instruction-following abilities, but maintain well-calibrated likelihood probabilities during the pre-training stage for all languages.

\subsection{Results regarding Different Languages on LA Tasks}
\label{ssec:la_lang}

\paragraph{Findings:}
\textbf{Linguistic dominance is manifested in \textit{English} with superior confidence estimation performances on LA tasks for multilingual LLMs.}
Prior works only validate the efficacy of prompt-based confidence estimations in \textit{English}.
Our findings indicate that the methods are also preferable in other languages and performances fluctuate in different languages.
In Table \ref{table:la_auroc_ece}, ARC. scores are less fluctuating across different languages while ECE in \textit{English} (23.04 and 22.77) performs better than in other languages on both {{GPT-3.5}} and {{Llama-3.1}}.
We also report the accuracy on LA datasets in Figure \ref{fig:la_accu}. \textit{English} consistently performs better across all datasets exclusively GSM8K.
Generally, prompting in \textit{English} outperforms others, hence responding in \textit{English} on LA tasks can be adequately credible and accurate where linguistic dominance is leading in \textit{English}.

\paragraph{Analysis:}
Despite the powerful multilingual capacity of LLMs, discrepancies exist in the quantity of distinct linguistic training corpora available for each language.
Results in Table \ref{table:la_auroc_ece} suggest that {ARC. is a metric not significantly related to language usage in LLMs, while the strong performance of ECE in \textit{English} can be attributed to the extensive training corpus or calibrations conducted during training in \textit{English}}.
As the only middle-resource language, \textit{Thai} exhibits a notably lower level of reliability compared to the other high-resource languages.

{Considering consistency check in Table \ref{table:check}, the lowest filter rate in GSM8K translation indicates that mathematical reasoning tasks are minimally affected by language bias.}
As a result, accuracy fluctuations across different languages on GSM8K are relatively small.
For that \textit{Chinese} exhibits slightly superior mathematical capabilities compared to English on GSM8K on both {{GPT-3.5}} and {{Llama-3.1}} (Accu. 48.09 and 63.96), it is hypothesized that pre-training corpora contain a substantial amount of Chinese mathematical problems.


\begin{figure*}[ht!]
    \centering
    \includegraphics[width=0.99\textwidth]{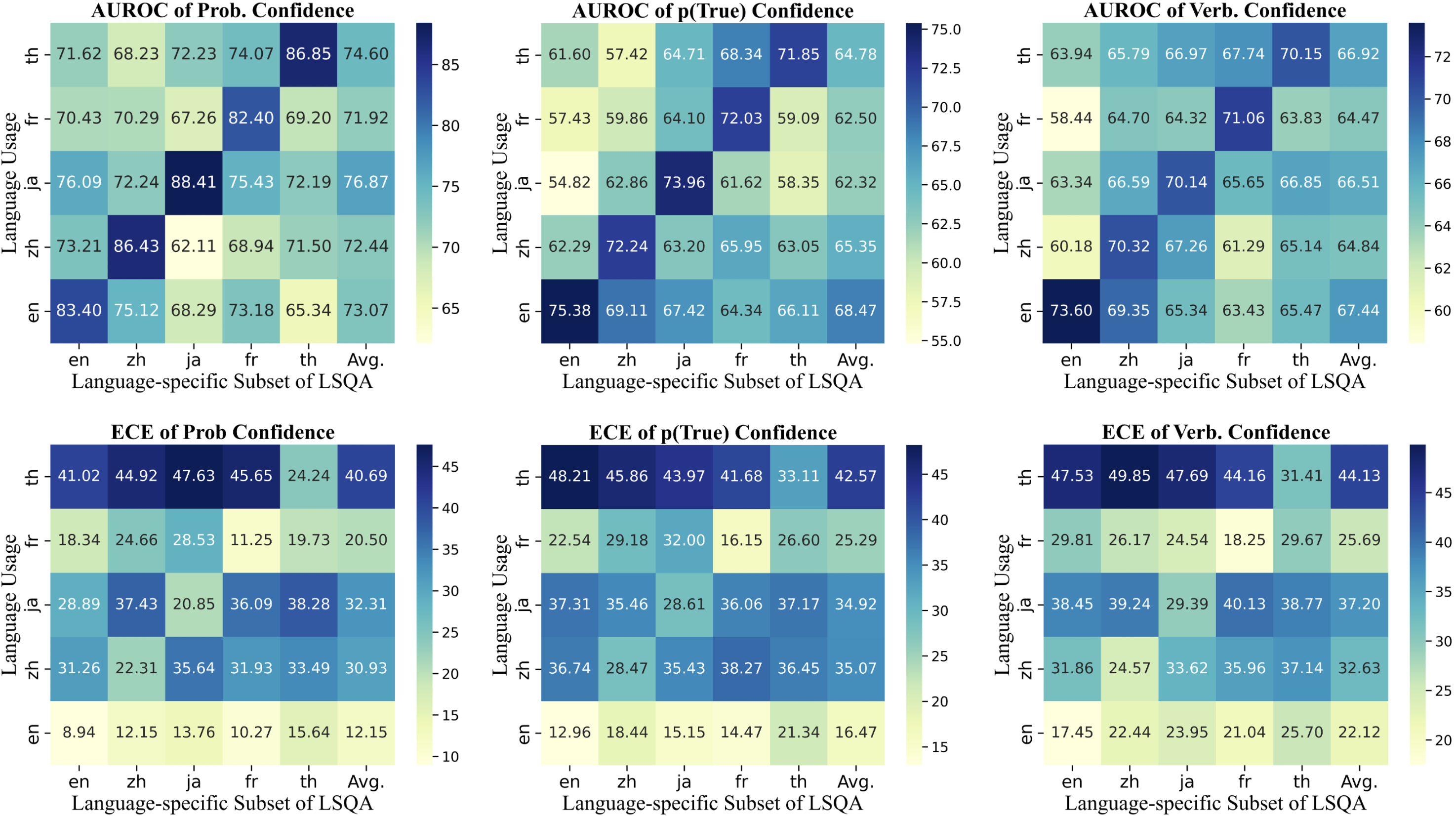}
    \caption{Experimental results of {AUROC} and {ECE} of three confidence estimation methods on five language-specific subset of {LSQA} using {{Llama-3.1}}.}
    \label{fig:ls_auroc_ece}
\end{figure*}

\begin{figure}[ht!]
    \centering
    \small
    \includegraphics[width=0.36\textwidth]{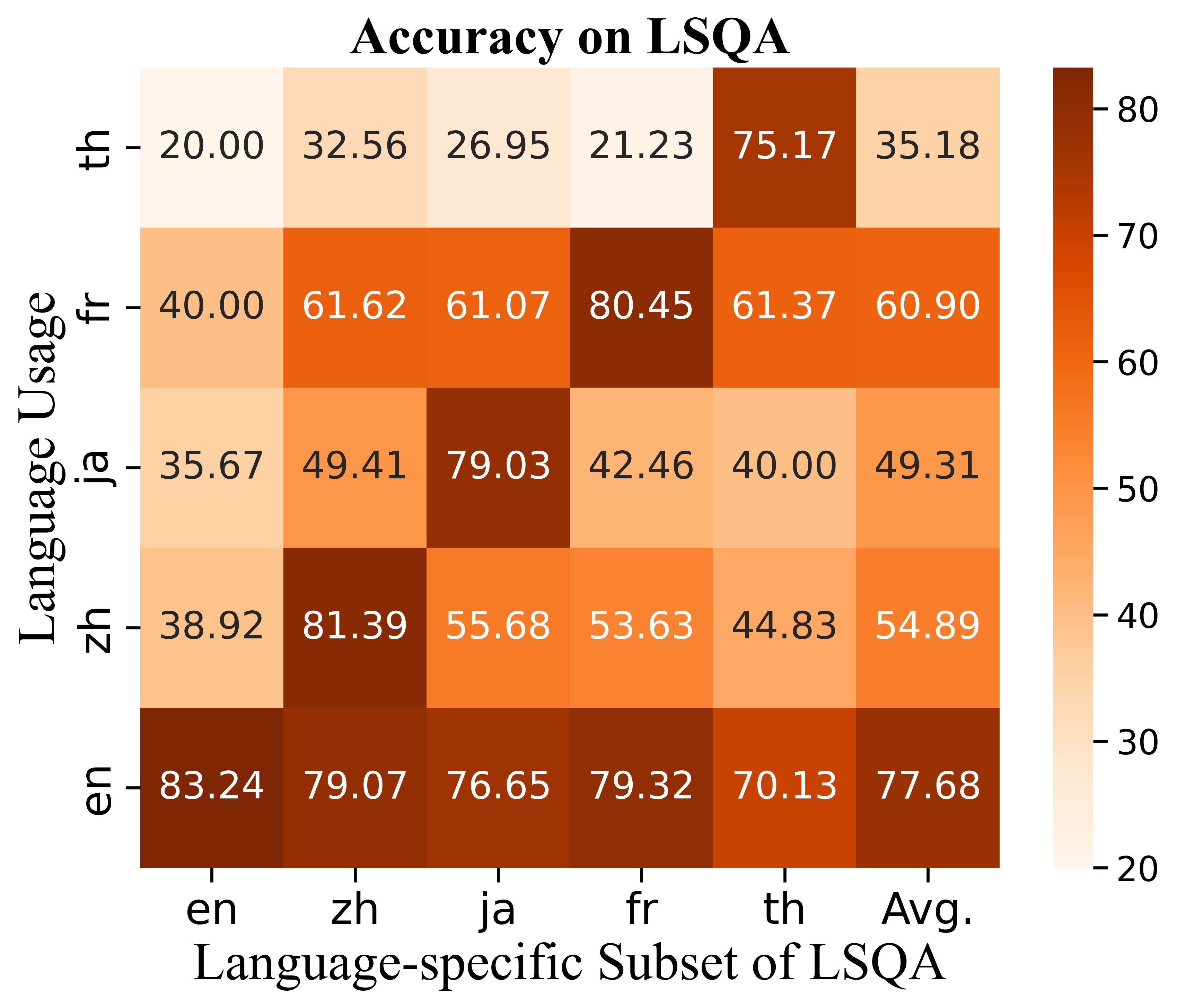}
    \caption{Experimental results of {Accuracy} on five language-specific {LSQA} subsets using {{Llama-3.1}}.}
    \label{fig:ls_accu}
\end{figure}

\section{Experiments on LS Task}

For the LS task, we present the confidence estimation results on five language-specific subsets of LSQA in Figure \ref{fig:ls_auroc_ece} and \ref{fig:ls_accu} in Sec. \ref{ssec:ls_res}.
Based on the findings, we then propose a \textbf{Native-Tone Prompting (NTP)} strategy to better leverage linguistic dominance to improve the LLMs' reliability and accuracy on LS task in Sec. \ref{ssec:ntp}.

\subsection{Results of Different Language-specific Subsets on LS Task}
\label{ssec:ls_res}


\paragraph{Findings:}
\textbf{Applying prompts in query-related language demonstrates linguistic dominance with better multilingual confidence estimations on LS task.}
In Figure \ref{fig:ls_auroc_ece}, the diagonal values of the ECE and ARC. heatmaps of {{Prob.}} are more pronounced, indicating that {when using {Prob.} confidence, linguistic dominance is more apparent compared to {{p(True)}} and {{Verb.}}}.
Consequently, we have opted {{Prob.}} for subsequent experiments in Sec. \ref{ssec:ntp}.
Additionally, in Figure \ref{fig:ls_accu}, although prompting in \textit{English} performs well and stable across different subsets, there is a noticeable improvement in accuracy when prompts are related to each subset's language.
In comparison with the LA tasks in Sec. \ref{ssec:la_res} where linguistic dominance is primarily manifested in \textit{English}, on LSQA, linguistic dominance is determined by specific language of the subset.

\paragraph{Analysis:}
The linguistic dominance on LSQA can be conjectured to stem from the fact that {such data pertaining to the language-specific cultural, geographical, or social contexts are already included in the pre-training corpora of their respective languages with higher certainty or confidence, thereby achieving optimal performance when prompting in their respective specific languages.

\subsection{Results of Native-Tone Prompting (NTP) Strategy on LS Task}
\label{ssec:ntp}



Confidence estimation performance differences caused by linguistic dominance phenomena on the LS task motivate us to explore the improving method.
Inspired by results in Sec. \ref{ssec:ls_res} on each language-specific LSQA subset, we propose a simple yet effective Native-Tone Prompting (NTP) strategy to achieve better confidence estimation performance on the LS task.
NTP first prompts LLMs to identify the language context of the question, and then uses that query-related language to answer the question, effectively exhibiting a ``native tone'' that is more familiar in that language-related context.
We present the results of prompting by any single language versus NTP on LSQA in Table \ref{table:ntp}.
The prompt of NTP is presented in the Appendix \ref{appendix:prompt}.

\begin{table}[ht!]
    \centering
    \footnotesize
    \resizebox{.48\textwidth}{!}{\begin{tabular}{c|cccccc}
    \toprule
      \multirow{1}{*}{\bf Prompt} & \bf {en} & \bf {zh} & \bf {ja} & \bf {fr} & \bf {th} & \bf NTP \\
        \hline
       \bf Accu. $\uparrow$ & 77.68 & 60.16 & 44.64 & 60.90 & 35.18 & \bf 79.46 \\
       \bf ARC. $\uparrow$ & 73.07 & 72.44 & 76.87 & 71.92 & 74.60 & \bf 77.25 \\
       \bf ECE $\downarrow$ & 12.15 & 30.93 & 32.31 & 20.50 & 40.69 & \bf 10.28 \\
    \bottomrule
    \end{tabular}}
    \caption{Experimental results of overall {Accu.}, {ARC.}, and {ECE} on the LSQA dataset by prompting using different languages and our proposed {NTP} method.}
\label{table:ntp}
\end{table}

\paragraph{Findings:}
\textbf{Prompting LLMs using the query-related language can enhance the reliability of confidence estimations and accuracy on LS tasks}, which provides an insight to improve LLMs' reliability regarding the prompt language usage.
In Table \ref{table:ntp}, the experiments demonstrate that NTP better leverages the inherent linguistic dominance, thereby yielding more reliable and accurate results than any single language prompt, validating the effectiveness of NTP on the LS task.

\paragraph{Analysis:}

Results in Table \ref{table:ntp} indicate that the multilingual capabilities and reliability of LLMs are still constrained by the imbalanced training corpus among diverse languages.
The reliability and accuracy on \textit{English}, serving as the primary training corpora, have not been adequately generalized to other languages.
Even for semantically equivalent queries in different languages, the reliability of responses cannot be consistently maintained.


\section{Discussion}

\paragraph{Extended Confidence Estimations}
In Appendix \ref{ssec:conf_estimation}, we further investigate three other confidence estimation methods including 1) paraphrasing the questions; 2) sampling multiple responses \citep{xiong2024can}; and 3) introducing Chain-of-Thought (CoT) \citep{jason2022chain} on both LS and LA tasks.
As presented in Table \ref{table:conf_method_la} and Figure \ref{fig:conf_method_ls} in the Appendix, \textbf{all findings on extended three confidence estimations are consistent with previous analysis across all languages}.
The questions after paraphrasing still maintain semantic equivalence without obvious perturbations for all languages, and LLMs are robust in multilingual confidence estimations to different questions with similar meanings.
p(True) and Verb. methods outperform sampling-based methods as the high temperature may incur variability in output spaces which undermines the reliability of QA tasks for all languages.
LLMs' CoT ability can be generalized to multilingual domains, thus benefiting multilingual confidence estimations.



\paragraph{Extended Languages}
In Appendix \ref{ssec:language}, we also extend the investigations on other five languages derived and translated from TriviaQA into \textit{Korean}, \textit{Arabic}, \textit{German}, \textit{Indonesian}, and \textit{Italian} as in Sec. \ref{sec:dataset}.
As in Table \ref{table:other_language} in the Appendix, linguistic dominance is still performed in \textit{English} than other languages on the LA task.
Low-resource languages demonstrate poor performance in ECE.
For the LS task, we also develop a small-size LSQA subset for the above five languages in Table \ref{table:ntp_other} to conduct the NTP method.
Experiments suggest that \textbf{NTP can also generalize and improve the reliability and accuracy in such middle- or low-resource languages}.

\section{Related Works}
\label{sec:related}

\paragraph{Confidence Estimation for LLMs}
Previous confidence estimation methods can be categorized into five classes, as illustrated in Figure \ref{fig:conf} and Appendix \ref{appendix:uncer}.
\textbf{\circledone~Probability-based}:
\citet{vazhentsev-etal-2023-efficient} intermediately quantifies sentence uncertainty over token probabilities;
\textbf{\circledtwo~$p(\mathrm{True})$-based}:
\citet{kadavath2022language} instructs the LLM to self-evaluate the correctness of the generated answer by directly accessing $p(\mathrm{True})$;
Both \circledone ~and \circledtwo ~require access to token probabilities and thus are limited to \textbf{white-box LLMs}.
\textbf{\circledthree~Self-verbalized}:
LLMs' remarkable instruction-following ability provides a view of expressing confidence in words \citep{lin2022teaching,zhou-etal-2023-navigating,tian-etal-2023-just,xiong2024can};
\textbf{\circledfour~Sampling-based}:
By sampling multiple responses to a given question, \citet{xiong2024can} aggregates all the confidence scores as the indicator.
\textbf{\circledfive~Training-based}:
\citet{lin2022teaching,kadavath2022language} propose to train an external module to improve confidence estimations.


\paragraph{Multlingual LLMs}

Most recent LLMs primarily pre-trained on English corpora have showcased remarkable capabilities \cite{pires-etal-2019-multilingual,shi2023language,openai2023gpt4}.
However, their efficacy in other low-resource languages remains limited.
Many research works have extended various tasks in multilingual domains such as claim fact-checking \citep{pikuliak-etal-2023-multilingual} and jailbreak problem \citep{deng2024multilingual}.
Prior studies have also explored diverse cross-lingual applications \cite{wang2023zeroshot,wang-etal-2023-towards-unifying,qin-etal-2022-gl}.

\section{Conclusion}
\label{sec:conclu}

This study underscores the necessity of advancing multilingual confidence estimation methods for LLMs to ensure their reliability across diverse linguistic contexts. 
The proposed \textsc{MlingConf} serves as a valuable and noteworthy benchmark to address the gap in multilingual confidence estimation research.
Our findings demonstrate the variability of multilingual confidence estimations on both LA and LS scenarios, revealing the influence of linguistic dominance on different tasks.
This leads to the NTP strategy, improving accuracy and reliability by aligning the response language with the linguistic context of the query for LS tasks.
These insights and the introduction of MlingConf datasets pave the way for future research, enhancing the global applicability and reliability of LLMs.


\section*{Limitations}
\label{sec:limit}

The limitations and prospects for future research are outlined as follows:

\paragraph{Expensive Costs to Obtain High-Quality Low-Resource Languages}
The present study is constrained by the substantial cost associated with the API cost using {\textbf{GPT-4}} for translation as well as linguistic verification.
This multilingual research is restricted to five languages in the first version.
This initial phase aims to delve into confidence estimation within multilingual domains. 
Our future endeavors will involve the expansion of the benchmark dataset, encompassing additional languages and datasets to enrich our investigations.

\paragraph{Native-Tone Prompting is a Primary Version}
Although the proposed Native-Tone Prompting method can enhance the accuracy and reliability of LS tasks, it still relies on external prompts to determine which language domain the query pertains to.
Moving forward, it is promising to broaden the scope of developing a cross-lingual method that can directly transfer the specific language dominance to other language contexts, thereby facilitating multilingual confidence estimation abilities for LLMs.

\section*{Ethics Statement}
\label{sec:ethics}

In this paper, we introduce several self-constructed multilingual datasets derived from the publicly available dataset.
The selection of investigated languages in this work depends on whether we can employ appropriate linguistic experts.
Most linguistic specialists are M.Phil. or Ph.D. students majoring in linguistics and others are from crowd-sourcing platforms.
We meticulously adhered to legal and ethical standards throughout the data collection process, prioritizing privacy and obtaining informed consent.
Linguistic experts were furnished with comprehensive details regarding the study's objectives, data collection methodologies, and associated risks or benefits.
They were afforded the opportunity to seek clarification and voluntarily provide consent before their involvement.
All collected data were solely utilized for research purposes.

\section*{Acknowledgements}
\label{sec:acknow}

This work was partially supported by Hong Kong RGC GRF No. 14206324, CUHK direct grant No. 4055209, and CUHK Knowledge Transfer Project Fund No. KPF23GWP20.

\bibstyle{acl_natbib}
\bibliography{anthology,custom}


\appendix

\section{Language Information}
\label{appendix:lang}

The basic information of ISO codes and the language family of the investigated languages is presented in Table \ref{table:lang}.
The investigated languages from widely spoken to lesser-known ones in this work are selected following three principles.

1) Following \citep{lai-etal-2023-chatgpt,deng2024multilingual} which determines the resource levels for each language by utilizing the data ratio from the CommonCrawl corpus \footnote{\href{http://commoncrawl.org/}{http://commoncrawl.org/}}, we select three languages (\textit{Chinese}, \textit{Japanese}, and \textit{French}) in high-resource category whose data ratio exceeds 1\%, and one language (\textit{Thai}) from medium-resource class that falls between 0.1\% and 1\%.
To ensure the confidence estimation ability can be observed, the low-resource languages less than 0.1\% are omitted and left for future works. 

2) This selection ensures coverage of a wide range of linguistic characteristics from different language families as in Table \ref{table:lang}.
A language family represents a collective of cognate languages stemming from a common ancestral source, serving as a focal point within the domain of linguistics \footnote{\href{https://en.wikipedia.org/wiki/Language_family}{https://en.wikipedia.org/wiki/Language\_family}}.

3) For each selected language, we can employ one linguistic expert for the human check to ensure the data quality;

\begin{table}[ht]
    \centering
    \footnotesize
    {\begin{tabular}{ccc}
    \toprule
    & {\bf{ISO 639-1}} & \multirow{1}{*}{\bf Family} \\
    \hline
    English & \bf {en} & Indo-European \\
    French & \bf {fr} & Indo-European \\
    Chinese & \bf {zh} & Sino-Tibetan \\
    Japenese & \bf {ja} & Japanese-Ryukyuan \\
    Thai & \bf {th} & Kra–Dai \\
    Indonesian & \bf {id} & Indo-European \\
    German & \bf {de} & Indo-European \\
    Arabic & \bf {ar} & Afro-Asiatic \\
    Korean & \bf {ko} & Koreanic \\
    Italian & \bf {it} & Indo-European \\
    \bottomrule
    \end{tabular}}
    \caption{List of International Standard Organization (ISO) 639-1 codes and language family information.}
\label{table:lang}
\end{table}


\section{Dataset Details}
\label{appendix:data}

\paragraph{TriviaQA}

The TriviaQA dataset \cite{joshi-etal-2017-triviaqa} is a realistic text-based reading comprehension question-answering dataset containing 650K question-answer-evidence triples from 95K documents collected from Wikipedia and the websites, served as a benchmark for evaluating machine comprehension and question-answering systems, which is more challenging than standard QA benchmark datasets where the answer spans can be directly retrieved and copied.

\paragraph{GSM8K}

GSM8K (Grade School Math 8K) \citep{cobbe2021training} is a dataset of 8.5K high quality linguistically diverse grade school math word problems.
The dataset was created to support the task of question answering on basic mathematical problems that require multi-step reasoning to solve.

\paragraph{CommonsenseQA}

CommonsenseQA \citep{talmor2019commonsense} is a new multiple-choice question answering dataset that requires different types of commonsense knowledge to predict the correct answers.
The dataset consists of 12,247 questions with 5 choices each.

\paragraph{SciQ}
The SciQ dataset \citep{welbl2017sciq} contains 13,679 crowdsourced science exam questions about Physics, Chemistry and Biology, among others. The questions are in multiple-choice format with 4 answer options each. For the majority of the questions, an additional paragraph with supporting evidence for the correct answer is provided.

\paragraph{LSQA}
We present two examples of the LSQA dataset in \textit{English}- and \textit{Japanese}- specific subsets in Figure \ref{fig:ls_examples}.


\section{Prompt Details}
\label{appendix:prompt}


The translation prompt for multilingual dataset construction and semantic equivalence comparison prompt for consistency check in Sec, \ref{sec:dataset} are presented in \ref{fig:prompt_trans} and \ref{fig:prompt_semantic} respectively.
Standard multilingual Question-Answering prompts are in \ref{fig:prompt_infer}.
Multilingual confidence estimations of \textbf{\textit{P(True)}} and \textbf{\textit{Verb.}} are presented in Fig. \ref{fig:prompt_ptrue} and \ref{fig:prompt_verb}.
Notably, the prompts for self-reflected true probability confidence estimation are followed by previous work \cite{kadavath2022language,kuhn2023semantic}.

\begin{figure*}[t]
    \centering
    \includegraphics[width=0.97\textwidth]{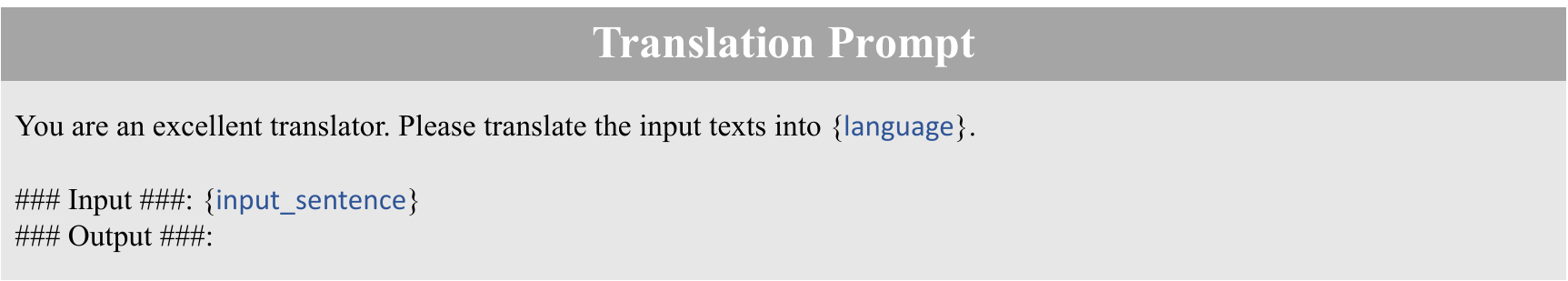}
    \caption{Translation prompt.}
    \label{fig:prompt_trans}
\end{figure*}

\begin{figure*}[t]
    \centering
    \includegraphics[width=0.97\textwidth]{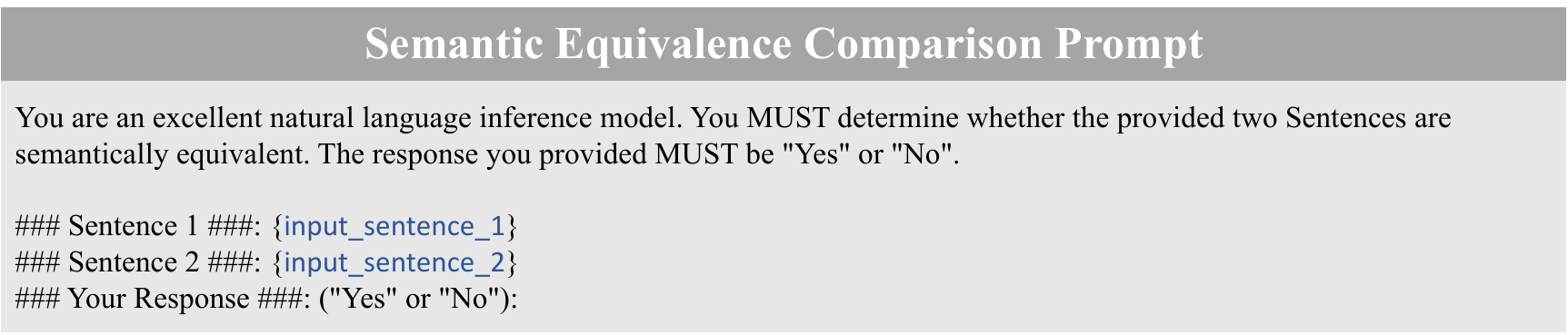}
    \caption{Semantic equivalence comparison prompt.}
    \label{fig:prompt_semantic}
\end{figure*}

\begin{figure*}[t]
    \centering
    \includegraphics[width=0.97\textwidth]{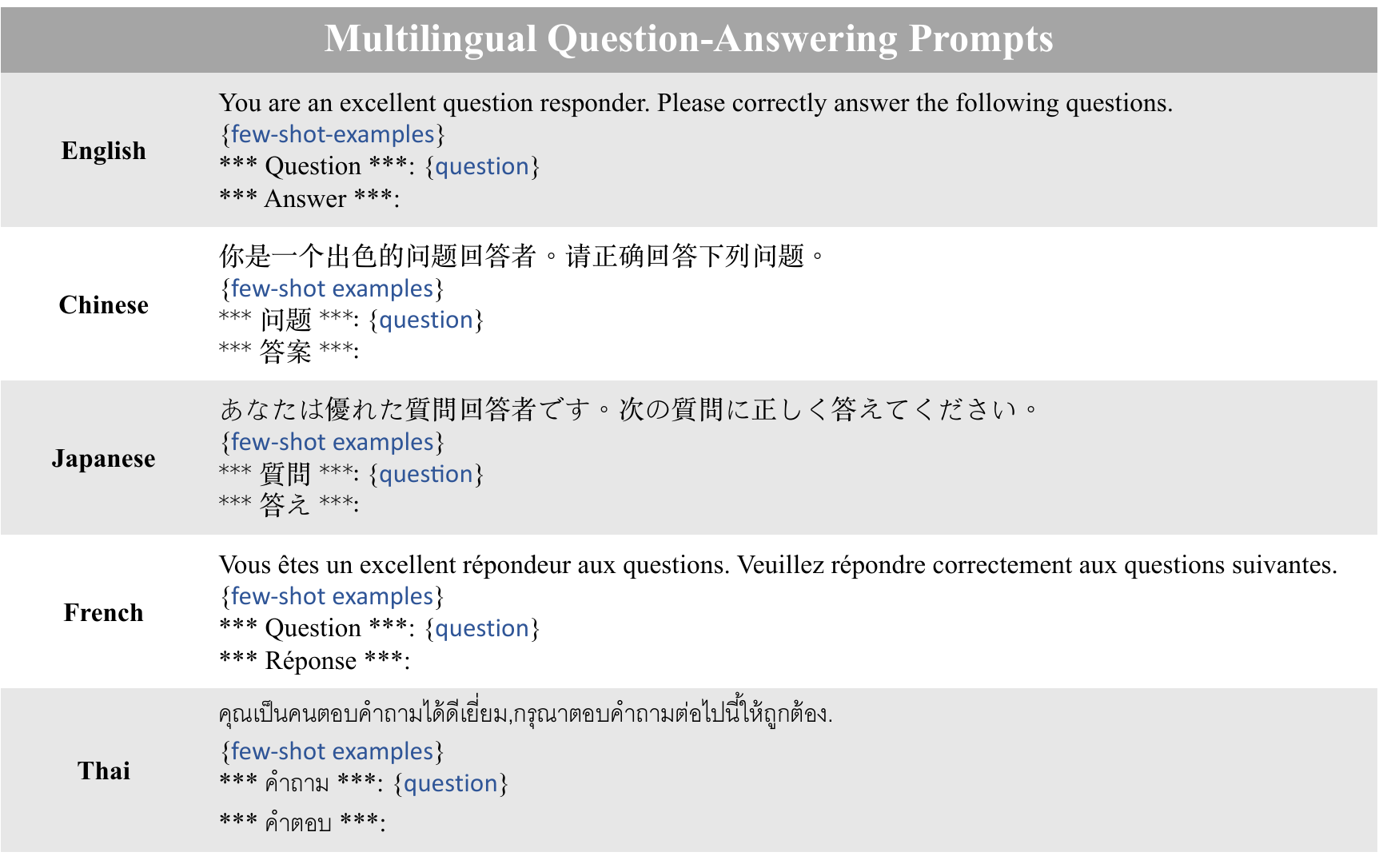}
    \caption{Multilingual Question-Answering prompts.}
    \label{fig:prompt_infer}
\end{figure*}

\begin{figure*}[t]
    \centering
    \includegraphics[width=0.97\textwidth]{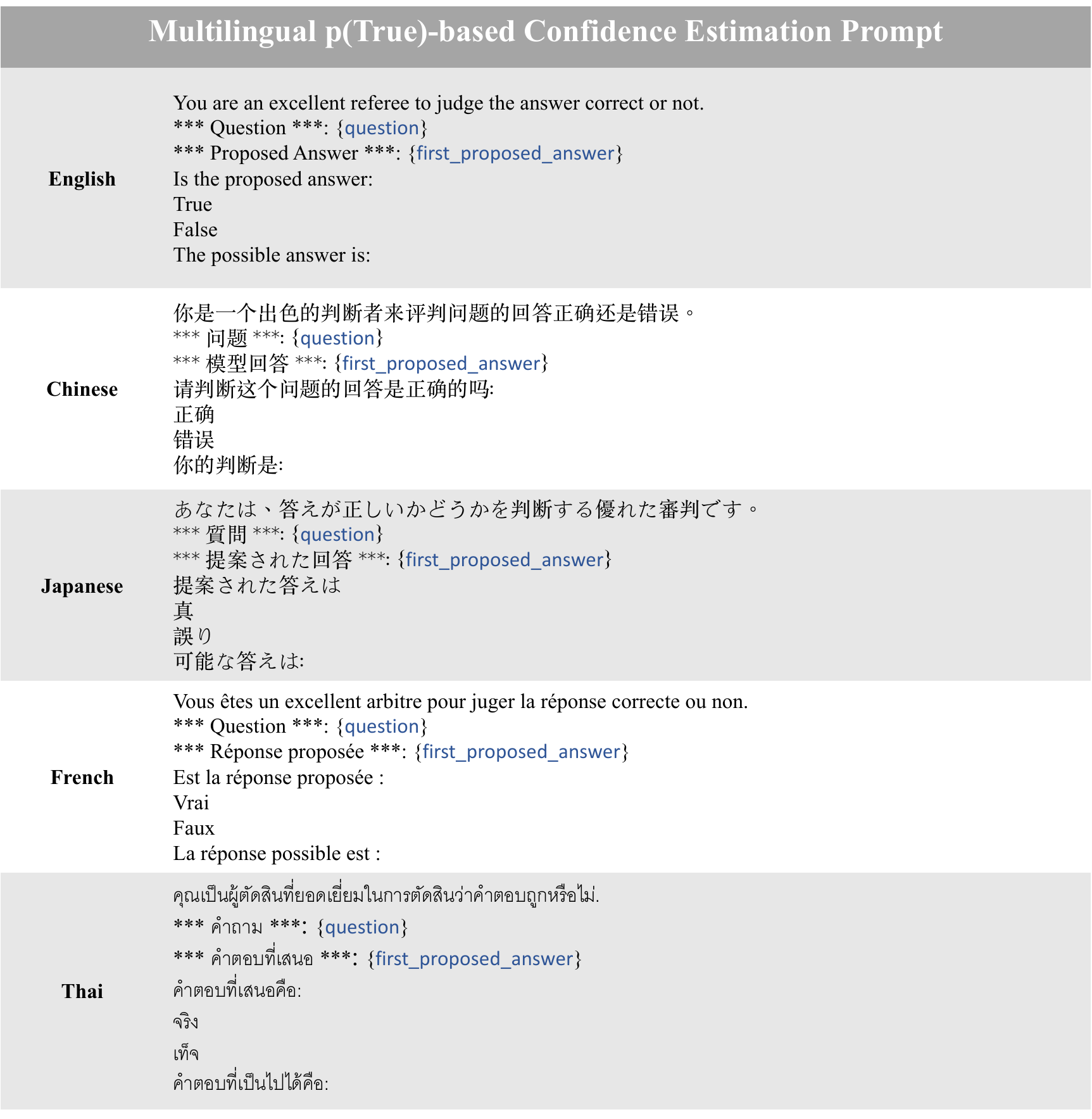}
    \caption{Multilingual \textit{\textbf{p(True)}}-based Confidence Estimation Prompt.}
    \label{fig:prompt_ptrue}
\end{figure*}

\begin{figure*}[t]
    \centering
    \includegraphics[width=0.97\textwidth]{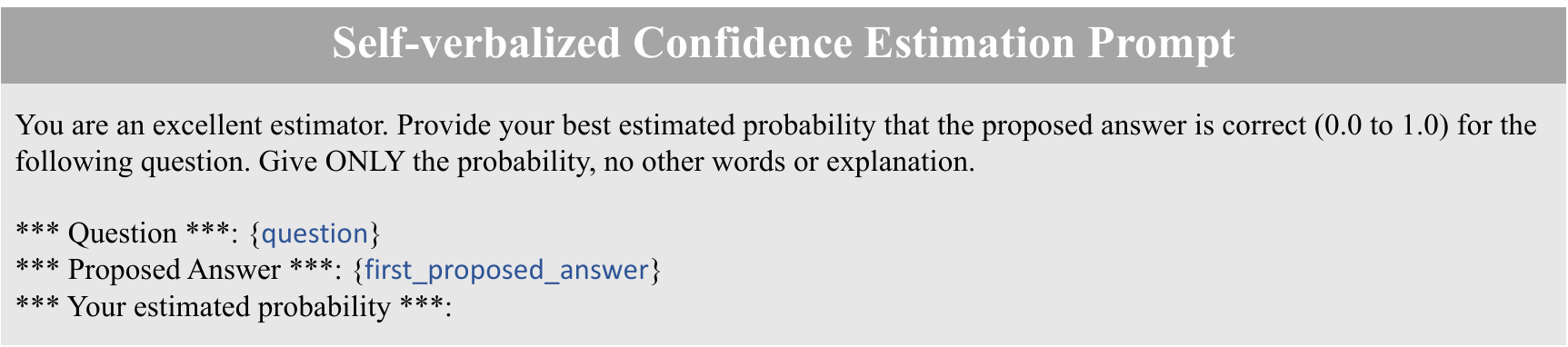}
    \caption{Self-verbalized confidence estimation prompt.}
    \label{fig:prompt_verb}
\end{figure*}

\begin{figure*}[t]
    \centering
    \includegraphics[width=0.97\textwidth]{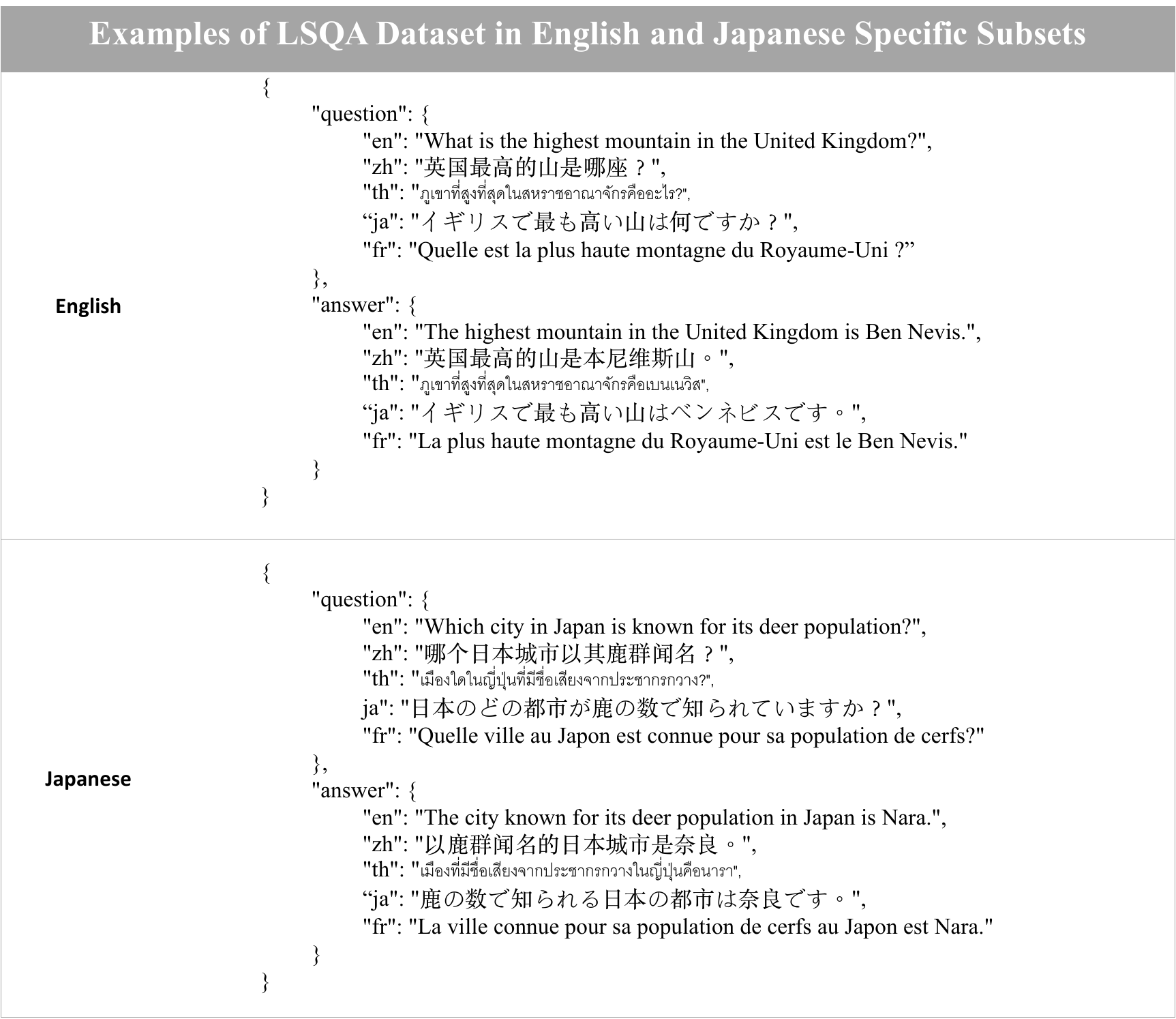}
    \caption{Examples of the LSQA ataset in \textit{English} and \textit{Japanese} specific subsets.}
    \label{fig:ls_examples}
\end{figure*}

\begin{figure*}[t]
    \centering
    \includegraphics[width=0.97\textwidth]{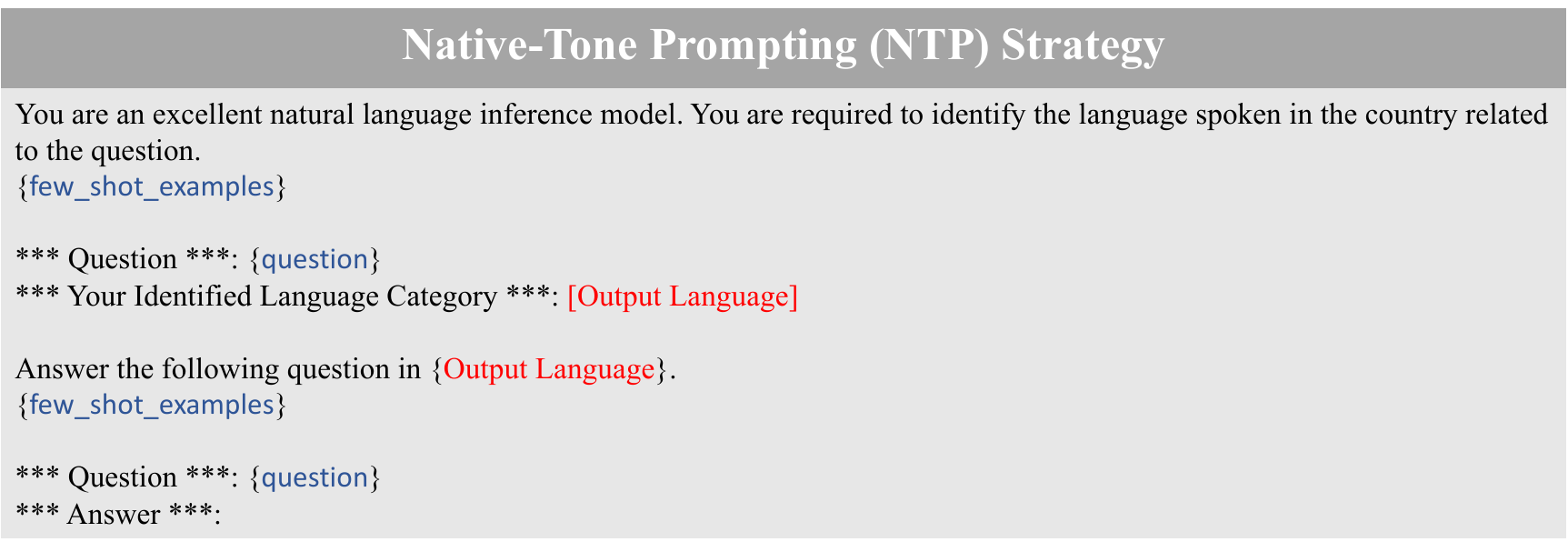}
    \caption{Native-tone prompting (NTP).}
    \label{fig:ntp}
\end{figure*}


\section{Metric Details}
\label{appendix:metric}

\paragraph{Expected Calibration Error (ECE)}

We partition the inference results into $M$ disjoint bins $\left\{B_m \right\}_{m=1}^M$ based on the confidence scores $\left\{q_i \right\}$, compute the average confidence score in $\left (\frac{m-1}{M},\frac{m}{M}\right ]$ for the $m$-th bin $B_m$, and compare it with the average true accuracy $\mathsf{acc}(B_m)$ of the answers within $B_m$. 
The ECE is calculated by:

\begin{align}
    \mathrm{ECE}=\sum_{m=1}^M\frac{|B_m|}{n}\left | \mathsf{acc}(B_m)-\mathsf{conf}(B_m) \right |
\end{align}

The average accuracy $\mathsf{acc}(B_m)$ and confidence $\mathsf{conf}(B_m)$ of the answers in $B_m$ is obtained by:

\begin{align}
    \mathsf{acc}(B_m)&=\frac{1}{|B_m|}\sum_{a_i\in _m}\mathbb{I}(\hat b_i=b_i) \\
    \mathsf{conf}(B_m)&=\frac{1}{|B_m|}\sum_{a_i\in _m}q_i
\end{align}
where $a_i$, $b_i$, $\hat b_i$, and $q_i$ indicate the input data, label, prediction result, and confidence score respectively for the $i$-th sample.

\paragraph{Accuracy} For closed-book QA evaluation, we observe that simply applying EM may misjudge the correct answers.
We compare several variants of EM as in Table \ref{table:em} and report their successful judgments on responses of 20 selected samples that are misjudged using EM, where PEM, RRM, and PREM indicate Positive-EM, Recall-EM, and Positive-Recall-EM and the mathematical explanations are presented in Table \ref{table:em}.
Upon human discrimination, EMPR exhibits the lowest failure rate and is therefore selected as the evaluation metric for this work.

\begin{table}[ht]
    \centering
    \footnotesize
    {\begin{tabular}{ccc}
    \toprule
    \bf Variant & \bf Explanation & \bf \# Fail \\
    \hline
    EM & $\boldsymbol y \equiv \boldsymbol {\hat y}$ & 20 \\
    PEM & $\boldsymbol y\in \boldsymbol {\hat y}$ & 16 \\
    REM & $\boldsymbol {\hat y} \in \boldsymbol y$. & 6 \\
    PREM & $\boldsymbol y\in \boldsymbol {\hat y} \vee \boldsymbol {\hat y} \in \boldsymbol y$. & 2 \\
    \bottomrule
    \end{tabular}}
    \caption{Number of failed judgments by human check for different EM variants.}
\label{table:em}
\end{table}

\section{Appendix Experiments}
\label{appendix:exp}

\subsection{Experiments on Llama-2 and Vicuna}

We present the experimental results on {{Llama-2-13B-Chat}} ({{Llama-2}}) \footnote{\href{https://huggingface.co/meta-llama/Llama-2-13b-chat}{https://huggingface.co/meta-llama/Llama-2-13b-chat}} \citep{touvron2023llama2}, and {{Vicuna-7B-v1.5}} ({{Vicuna-v1.5}}) \footnote{\href{https://huggingface.co/lmsys/vicuna-7b-v1.5}{https://huggingface.co/lmsys/vicuna-7b-v1.5}} \citep{zheng2023judging} in Table \ref{table:append_exp}.
Results suggest that confidence estimation abilities are relatively weak in both {{Llama-2}} and {Vicuna-1.5} across three methods.

\subsection{Experiments of Extended Confidence Estimations}
\label{ssec:conf_estimation}

\subsubsection{Experiments of Multilingual Confidence Estimations with Paraphrasing}

Following \citet{xiong2024can}, we investigate the prompt sensitivity for multilingual confidence estimation by introducing perturbations in the questions. 
We utilize GPT-3.5 to paraphrase the questions in different ways to generate different responses.
We sample 200 questions from SciQ and prompt GPT-3.5 to paraphrase these questions. 
We also employ GPT-3.5 to check the semantic equivalence before and after paraphrasing to ensure the meaning is not changed. 
The AUROC and ECE results are presented in Table \ref{table:conf_method_la} and Figure \ref{fig:conf_method_ls}.
The findings and analysis are in Sec. \ref{ssec:la_res}.

\subsubsection{Experiments of Multilingual Confidence Estimations of Sampling}

To make comparisons, we also present the AUROC and ECE results of sampling-based confidence estimation methods on 200 samples from our multilingual SciQ datasets by setting Temperature T=0.8 on GPT-3.5. 
We cluster the sampled responses in semantic spaces and calculate the consistency score as \citet{xiong2024can} to represent the confidence.
As presented in Table \ref{table:conf_method_la} and \ref{fig:conf_method_ls}, the results demonstrate that our employed p(True) and Verb. methods outperform sampling-based methods as the high temperature may incur variability in output spaces which undermines the reliability of QA tasks.

\subsubsection{Experiments of Multilingual Confidence Estimations using CoT}

We supply the Chain-of-Thought (CoT) \citep{jason2022chain} for prompt-based confidence estimations of p(True) and Verbalized methods as in Table \ref{table:conf_method_la} and Figure \ref{table:conf_method_la}.
We present the AUROC and ECE results of p(True) and Verb. using CoT on 400 samples from SciQ and LSQA on GPT-3.5.
Results suggest that CoT can marginally enhance the reliability of prompt-based confidence estimations in various languages.

\begin{table}[ht!]
    \centering
    \footnotesize
    \resizebox{.44\textwidth}{!}{\begin{tabular}{c|ccccc}
    \toprule
      \multirow{1}{*}{\bf Lang.} & \bf {ko} & \bf {id} & \bf {it} & \bf {ar} & \bf {de} \\
        \hline
        \rowcolor{platinum}
        \multicolumn{6}{c}{\textbf{Prompt in \textit{English}}} \\
        \hline
       \bf Accu. $\uparrow$ & 24.39 & 40.60 & 34.58 & 22.64 & 54.78 \\
       \bf ARC. $\uparrow$ & 72.40 & 70.12 & 75.45 & 68.22 & 76.18 \\
       \bf ECE $\downarrow$ & 33.55 & 36.78 & 33.16 & 46.78 & 27.14 \\
        \hline
        \rowcolor{platinum}
        \multicolumn{6}{c}{\textbf{NTP Method}} \\
       \bf Accu. $\uparrow$ & 28.60 & 46.54 & 39.20 & 27.44 & 59.65 \\
       \bf ARC. $\uparrow$ & 74.66 & 78.52 & 77.23 & 70.17 & 79.60 \\
       \bf ECE $\downarrow$ & 28.10 & 32.44 & 30.50 & 42.76 & 23.18 \\
        \hline
    \bottomrule
    \end{tabular}}
    \caption{Experimental results of overall {Accu.}, {ARC.}, and {ECE} on the LSQA dataset by prompting using \textit{English} and {NTP} method on other five investigated languages.}
\label{table:ntp_other}
\end{table}

\begin{table*}[!ht]
    \centering
    \footnotesize
    \resizebox{.86\textwidth}{!}{\begin{tabular}{c|cc|cc|cc|cc|cc}
    \toprule
        \multirow{2}{*}{\bf Conf.} & \multicolumn{2}{c|}{\bf {en}} & \multicolumn{2}{c|}{\bf {zh}} & \multicolumn{2}{c|}{\bf {ja}} & \multicolumn{2}{c|}{\bf {fr}} & \multicolumn{2}{c}{\bf {th}} \\
         & \bf ARC. $\uparrow$ & \bf ECE $\downarrow$ & \bf ARC. $\uparrow$ & \bf ECE $\downarrow$ & \bf ARC. $\uparrow$ & \bf ECE $\downarrow$ & \bf ARC. $\uparrow$ & \bf ECE $\downarrow$ & \bf ARC. $\uparrow$ & \bf ECE $\downarrow$ \\
        \hline
        \hline
        \rowcolor{platinum}
        \multicolumn{11}{c}{\textbf{TVQA on \textsl{Llama-2}}} \\
        \hline
        \textbf{\textit{Prob.}} & 51.92 & 20.56 & 51.24 & 34.12 & 51.92 & 32.04 & 49.51 & 29.33 & 49.63 & 48.72 \\
        \textbf{\textit{p(True)}} & 55.89 & 17.09 & 82.65 & 46.11 & 81.76 & 43.47 & 65.79 & 29.59 & 70.62 & 55.99 \\
        \textbf{\textit{Verb.}} & 59.78 & 21.10 & 53.20 & 45.71 & 51.95 & 39.33 & 61.66 & 38.98 & 54.23 & 59.80 \\
        \hline
        \rowcolor{platinum}
        \multicolumn{11}{c}{\textbf{GSM8K on \textsl{Llama-2}}} \\
        \hline
        \textbf{\textit{Prob.}} & 42.72 & 32.18 & 50.46 & 33.59 & 50.35 & 51.67 & 43.60 & 36.25 & 55.11 & 52.87 \\
        \textbf{\textit{p(True)}} & 60.82 & 49.30 & 59.86 & 58.87 & 62.89 & 67.39 & 59.51 & 62.36 & 47.91 & 75.22 \\
        \textbf{\textit{Verb.}} & 59.39 & 43.65 & 53.29 & 54.59 & 53.40 & 49.27 & 53.26 & 37.61 & 54.53 & 56.96 \\
        \hline
        \rowcolor{platinum}
        \multicolumn{11}{c}{\textbf{CSQA on \textsl{Llama-2}}} \\
        \hline
        \textbf{\textit{Prob.}} & 49.30 & 30.40 & 49.95 & 31.72 & 50.28 & 43.28 & 49.72 & 27.40 & 50.23 & 40.84 \\
        \textbf{\textit{p(True)}} & 56.53 & 26.05 & 55.34 & 45.65 & 53.46 & 46.01 & 59.76 & 25.49 & 50.21 & 63.09 \\
        \textbf{\textit{Verb.}} & 53.64 & 19.54 & 51.74 & 24.06 & 50.36 & 34.03 & 52.93 & 15.08 & 50.73 & 62.01 \\
        \hline
        \rowcolor{platinum}
        \multicolumn{11}{c}{\textbf{SciQ on \textsl{Llama-2}}} \\
        \hline
        \textbf{\textit{Prob.}} & 55.40 & 24.65 & 76.39 & 44.42 & 74.97 & 45.56 & 62.32 & 39.76 & 51.93 & 59.05 \\
        \textbf{\textit{p(True)}} & 48.60 & 32.18 & 52.02 & 40.44 & 51.60 & 30.19 & 49.53 & 32.50 & 45.26 & 43.75 \\
        \textbf{\textit{Verb.}} & 56.34 & 19.89 & 55.17 & 41.36 & 55.58 & 37.20 & 60.27 & 39.14 & 71.17 & 54.95 \\
        \hline
        \rowcolor{platinum}
        \multicolumn{11}{c}{\textbf{TVQA on \textsl{Vicuna-1.5}}} \\
        \hline
        \textbf{\textit{Prob.}} & 45.45 & 35.34 & 48.43 & 47.07 & 51.75 & 36.63 & 46.13 & 35.19 & 53.17 & 40.73 \\
        \textbf{\textit{p(True)}} & 47.45 & 23.58 & 78.96 & 42.86 & 79.71 & 42.40 & 60.38 & 28.89 & 76.58 & 53.43 \\
        \textbf{\textit{Verb.}} & 55.74 & 21.41 & 52.98 & 57.36 & 50.94 & 54.89 & 55.46 & 42.76 & 45.86 & 71.83 \\
        \hline
        \rowcolor{platinum}
        \multicolumn{11}{c}{\textbf{GSM8K on \textsl{Vicuna-1.5}}} \\
        \hline
        \textbf{\textit{Prob.}} & 50.90 & 53.91 & 51.07 & 49.00 & 50.51 & 53.73 & 50.42 & 49.08 & 50.19 & 55.40 \\
        \textbf{\textit{p(True)}} & 65.40 & 68.30 & 67.28 & 59.33 & 51.09 & 60.70 & 66.78 & 55.69 & 52.86 & 60.83 \\
        \textbf{\textit{Verb.}} & 55.66 & 46.26 & 53.90 & 48.75 & 54.60 & 48.03 & 53.70 & 45.62 & 61.81 & 51.87 \\
        \hline
        \rowcolor{platinum}
        \multicolumn{11}{c}{\textbf{CSQA on \textsl{Vicuna-1.5}}} \\
        \hline
        \textbf{\textit{Prob.}} & 48.88 & 26.04 & 50.01 & 43.65 & 49.67 & 45.64 & 45.94 & 31.53 & 49.78 & 51.78 \\
        \textbf{\textit{p(True)}} & 65.00 & 27.06 & 57.39 & 35.80 & 57.62 & 38.54 & 48.21 & 25.95 & 50.53 & 55.54 \\
        \textbf{\textit{Verb.}} & 52.32 & 29.80 & 52.29 & 38.59 & 51.08 & 44.49 & 58.68 & 35.15 & 51.90 & 61.77 \\
        \hline
        \rowcolor{platinum}
        \multicolumn{11}{c}{\textbf{SciQ on \textsl{Vicuna-1.5}}} \\
        \hline
        \textbf{\textit{Prob.}} & 38.10 & 50.94 & 48.69 & 44.44 & 50.07 & 42.19 & 38.65 & 37.26 & 49.75 & 49.70 \\
        \textbf{\textit{p(True)}} & 45.78 & 31.29 & 73.55 & 40.17 & 66.85 & 45.15 & 59.20 & 42.18 & 74.16 & 58.34 \\
        \textbf{\textit{Verb.}} & 55.13 & 36.47 & 51.66 & 55.27 & 51.92 & 56.98 & 56.33 & 57.93 & 52.89 & 57.74 \\
        \hline
        \rowcolor{platinum}
        \multicolumn{11}{c}{\textbf{TVQA on \textsl{GPT-4o}}} \\
        \hline
        \textbf{\textit{Prob.}} & 74.22	& 11.38 & 72.00 & 22.34 & 73.45 & 20.17 & 76.18 & 12.43 & 78.33 & 27.48 \\
        \textbf{\textit{p(True)}} & 78.15 & 8.34 & 77.44 & 18.24 & 82.73 & 16.76 & 77.24 & 11.82 & 83.44 & 25.15 \\
        \textbf{\textit{Verb.}} & 79.33 & 8.63 & 78.16 & 16.33 & 81.56 & 17.18 & 79.68 & 10.37 & 86.08 & 26.44 \\
        \hline
        \rowcolor{platinum}
        \multicolumn{11}{c}{\textbf{GSM8K on \textsl{GPT-4o}}} \\
        \hline
        \textbf{\textit{Prob.}} & 61.50 & 21.37 & 67.34 & 24.31 & 63.55 & 26.47 & 62.45 & 28.15 & 60.44 & 31.25 \\
        \textbf{\textit{p(True)}} & 71.49 & 18.13 & 75.19 & 22.97 & 73.50 & 22.77 & 73.11 & 21.45 & 70.60 & 28.11 \\
        \textbf{\textit{Verb.}} & 75.28 & 16.55 & 74.30 & 21.48 & 75.16 & 21.64 & 71.65 & 23.79 & 67.54 & 27.43 \\
        \hline
        \rowcolor{platinum}
        \multicolumn{11}{c}{\textbf{CSQA on \textsl{GPT-4o}}} \\
        \hline
        \textbf{\textit{Prob.}} & 62.76 & 21.34 & 59.46 & 33.62 & 58.47 & 31.78 & 66.14 & 28.40 & 59.19 & 38.66 \\
        \textbf{\textit{p(True)}} & 67.40 & 17.56 & 64.75 & 28.97 & 63.89 & 21.40 & 72.18 & 19.70 & 68.56 & 31.65 \\
        \textbf{\textit{Verb.}} & 69.14 & 16.20 & 66.13 & 27.33 & 66.60 & 23.65 & 71.79 & 19.44 & 64.33 & 33.98 \\
        \hline
        \rowcolor{platinum}
        \multicolumn{11}{c}{\textbf{SciQ on \textsl{GPT-4o}}} \\
        \hline
        \textbf{\textit{Prob.}} & 73.20 & 27.24 & 76.44 & 33.37 & 79.23 & 39.16 & 74.81 & 31.70 & 78.20 & 44.60 \\
        \textbf{\textit{p(True)}} & 77.54 & 16.78 & 78.62 & 28.07 & 83.45 & 28.07 & 79.13 & 21.34 & 81.39 & 31.40 \\
        \textbf{\textit{Verb.}} & 78.18 & 17.90 & 79.13 & 25.16 & 84.52 & 25.44 & 78.63 & 20.56 & 79.27 & 33.22 \\
        \hline
    \bottomrule
    \end{tabular}}
    \caption{Experimental results of AUROC ({ARC.}) and {ECE} of three confidence estimation methods on four LA datasets on {\textsl{Llama-2}} and {\textsl{Vicuna-1.5}}.}
\label{table:append_exp}
\end{table*}


\begin{figure*}[ht!]
    \centering
    \includegraphics[width=0.99\textwidth]{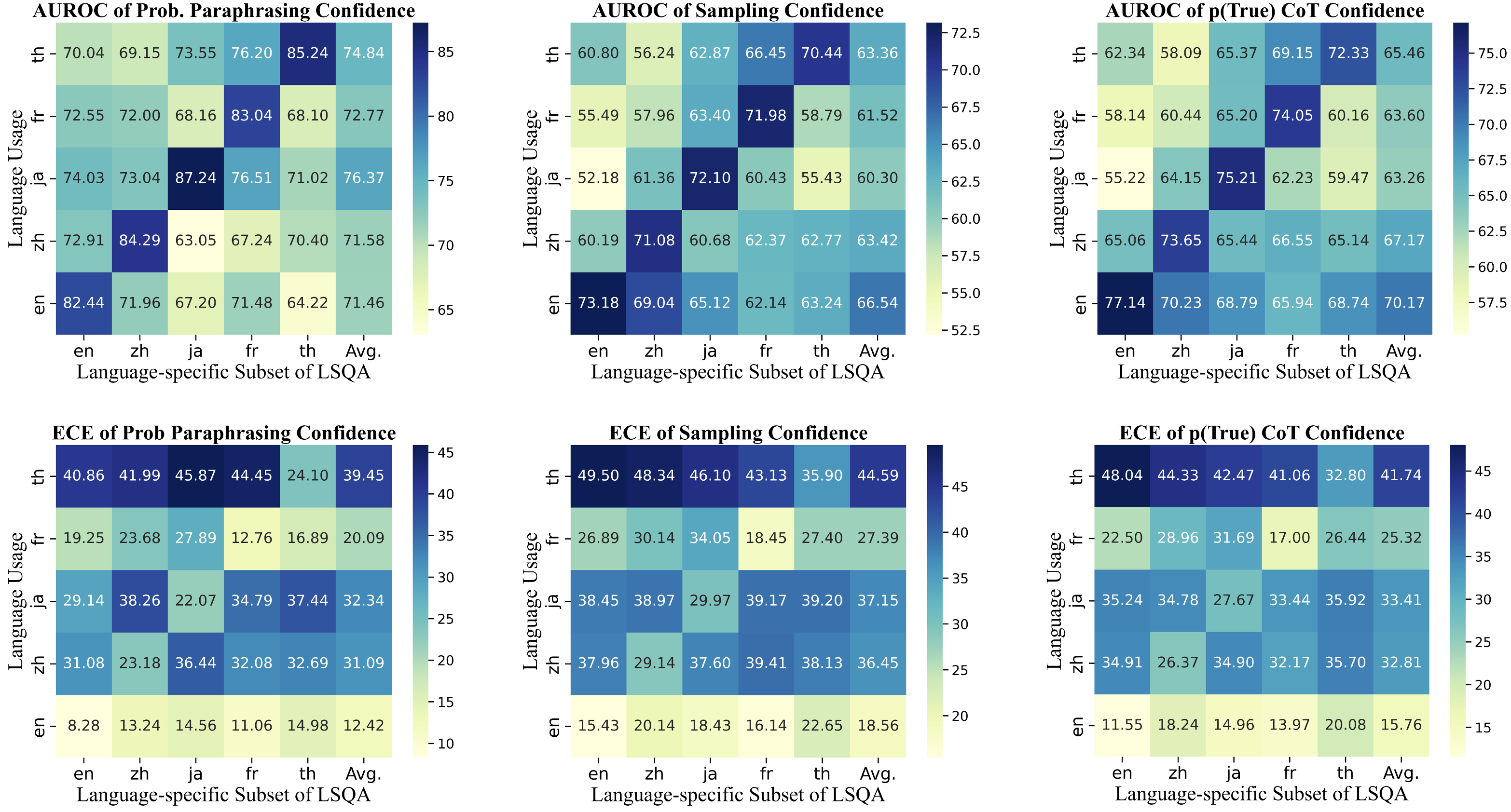}
    \caption{Experimental results of {AUROC} and {ECE} of three confidence estimation variants of paraphrasing, sampling, and CoT on LSQA for LS task on GPT-3.5.}
    \label{fig:conf_method_ls}
\end{figure*}

\begin{figure*}[ht!]
    \centering
    \includegraphics[width=0.66\textwidth]{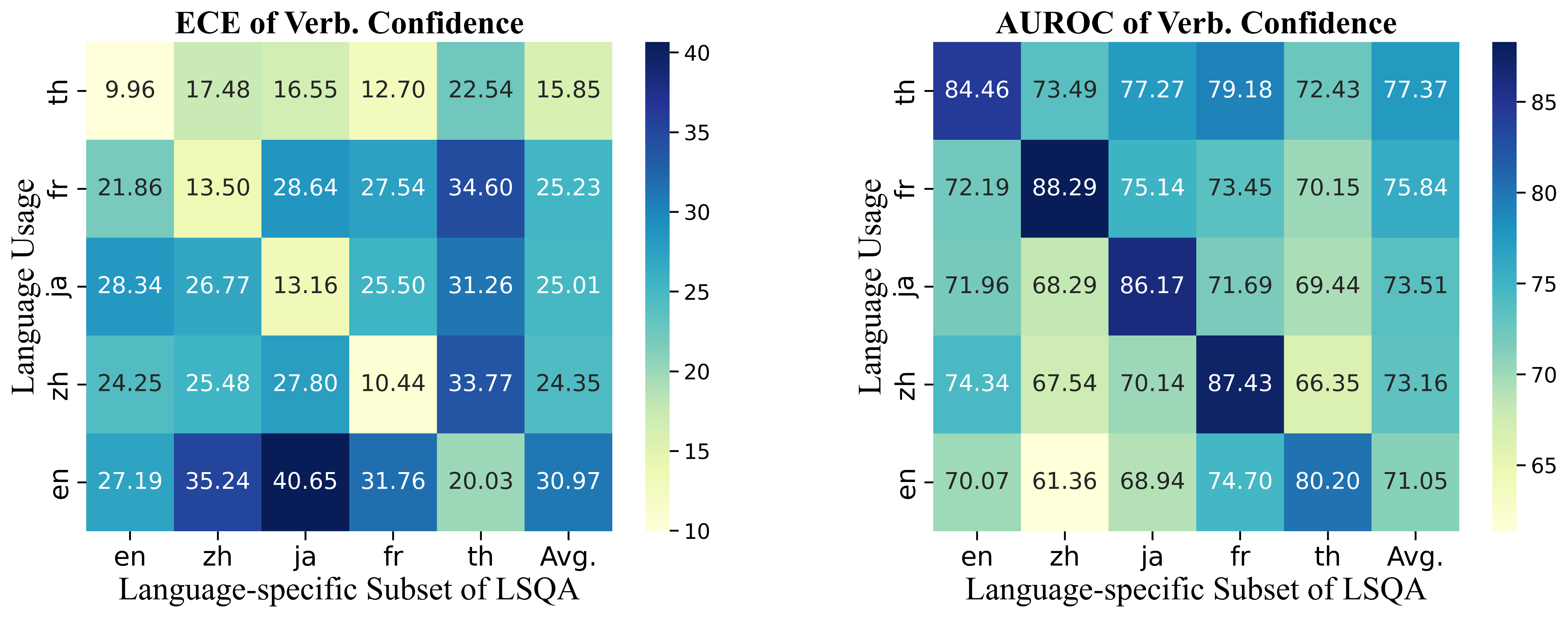}
    \caption{Experimental results of {AUROC} and {ECE} of verbalized confidence estimation on LSQA for LS task on GPT-4o.}
    \label{fig:conf_method_ls_4o}
\end{figure*}

\subsection{Experiments on Extended Languages}
\label{ssec:language}
To further validate the observed linguistic dominance in multilingual confidence estimations, we employ five subsets derived and translated from TriviaQA into \textit{Korean} (\texttt{\bf ko}), \textit{Arabic} (\texttt{\bf ar}), \textit{German} (\texttt{\bf de}), \textit{Indonesian} (\texttt{\bf id}), and \textit{Italian} (\texttt{\bf it}) as in Sec. \ref{sec:dataset}.
The LA experiments are conducted on dataset translated from TriviaQA in all investigated languages in Table \ref{table:other_language}.
We also develop small-size LSQA subsets for such languages and conduct LS experiments in \ref{table:ntp_other}.

\begin{table*}[ht!]
    \centering
    \footnotesize
    \resizebox{.93\textwidth}{!}{\begin{tabular}{c|cc|cc|cc|cc|cc}
    \toprule
        \multirow{2}{*}{\bf Conf.} & \multicolumn{2}{c|}{\bf {en}} & \multicolumn{2}{c|}{\bf {zh}} & \multicolumn{2}{c|}{\bf {ja}} & \multicolumn{2}{c|}{\bf {fr}} & \multicolumn{2}{c}{\bf {th}} \\
         & \bf ARC. $\uparrow$ & \bf ECE $\downarrow$ & \bf ARC. $\uparrow$ & \bf ECE $\downarrow$ & \bf ARC. $\uparrow$ & \bf ECE $\downarrow$ & \bf ARC. $\uparrow$ & \bf ECE $\downarrow$ & \bf ARC. $\uparrow$ & \bf ECE $\downarrow$ \\
        \hline
        \rowcolor{platinum}
        \multicolumn{11}{c}{\textbf{{SciQ} on \textsl{GPT-3.5}}} \\
        \hline
       \textbf{\textit{Prob.}} & 69.58 & 30.04 & 67.14	& 36.77 & 81.44 & 45.40 & 74.35 & 36.98 & 72.55 &	51.34 \\
       \textbf{\textit{p(True)}} & 72.80 & 23.86 & 77.56 & 31.99 & 82.44 & 38.27 & 72.00 & 40.13 & 63.45 & 40.80 \\
       \textbf{\textit{Verb.}} & 71.43 & 22.18 & 72.50 & 36.47 & 72.95 & 31.43 & 74.16 & 31.97 &	73.40 & 42.34 \\
       \hline
       \textbf{\textit{Re-Prob.}} & 67.47 & 28.16 &	72.86 & 33.43 & 75.69 & 41.05 & 71.40 & 34.88 & 80.37 & 48.96 \\
       \textbf{\textit{Re-p(True)}} & 74.14 & 25.14 & 82.66 & 32.04 & 76.96 & 36.70 & 71.48 & 42.13 & 64.44 & 42.05 \\
       \textbf{\textit{Re-Verb.}} & 73.80 & 21.96 & 73.40 & 35.13 & 79.49 & 30.60 & 66.16 & 32.65 & 73.19 & 40.44 \\
       \hline
       \textbf{\textit{Sampling}} & 67.55 & 27.40 & 71.69 & 37.97 & 74.07 & 42.09 & 67.94 & 40.04 & 66.50 & 48.65 \\
       \hline
       \textbf{\textit{CoT-p(True)}} & 73.65 & 22.95 & 80.05 & 29.90 & 82.16 & 37.10 & 71.92 & 30.86 & 65.90 &	40.19 \\
       \textbf{\textit{CoT-Verb.}} & 73.64 & 20.60 & 75.73 & 32.79 & 74.61 & 27.50 & 72.62 & 31.26 & 74.96 & 40.33 \\
    \bottomrule
    \end{tabular}}
    \caption{Experimental results of AUROC and ECE of several confidence estimation variants of paraphrasing the questions, sampling multiple responses, and adding CoT on SciQ for LA task on GPT-3.5.}
\label{table:conf_method_la}
\end{table*}

\subsection{Formatting P(True) Method}

The output format issue of the two prompt-based confidence estimation methods is not the primary focus of this study, nor have we observed previous works addressing this problem.
However, these issues posed significant challenges during our experiments. With strong instruction-following ability, GPT-3.5 typically generates outputs in the correct format.
In cases of occasional formatting errors in the outputs, we employed temperature sampling to re-generate the outputs until the correct format was achieved.

The other three LLMs—Llama-3.1-Instruct, Llama-2-Chat, and Vicuna-v1.5—also demonstrated relatively good adherence to instructions for verbalized confidence estimation.
If the correct format was not generated on the first attempt, we also employ temperature sampling multiple times to obtain the expected output.

For the P(True) method, however, output format discrepancies were more pronounced.
We explored two approaches: \textbf{rule-based post-processing} and \textbf{few-shot formatting}.
Initially, we attempted rule-based post-processing, where the ideal output format would directly consist of ``true'' or ``false'' following the input.
However, in practice, the models often included their own analysis, and in many cases, the generated sequence did not begin with the desired result.
To address this, we detected multilingual keywords within the generated sequence as in Fig. \ref{fig:prompt_ptrue}.

Since some keywords span more than one token in some languages, we first stored the tokenized sequence before decoding, along with the corresponding logits. 
We then extracted the logits associated with the tokenized keywords with the normalized or the first-token probability.
Despite these efforts, this approach still failed to consistently produce the expected outputs.

We subsequently turned to a few-shot approach.
To avoid biases from the order or quantity of ``true'' and ``false'' examples in the few-shot samples, we set the number of examples to 10, evenly split between ``true'' and ``false'' (five each), and randomized their order in every instance.
Ultimately, we found that the few-shot approach not only produced more stable output formats but also yielded more reliable AUROC and ECE results. Therefore, we adopted the few-shot method as our final approach.

Additionally, we considered a training-based method, where negative samples would be constructed to train a classifier head specifically designed to output ``true'' or ``false''.
However, this approach was prohibitively costly, as it would require training a separate head for each model in every language. Consequently, we decided against pursuing this method.

\begin{table*}[t!]
    \centering
    \footnotesize
    \resizebox{.99\textwidth}{!}{\begin{tabular}{c|cc|cc|cc|cc|cc|cc}
    \toprule
        \multirow{2}{*}{\bf Conf.} & \multicolumn{2}{c|}{\bf {en}} & \multicolumn{2}{c|}{\bf {ko}} & \multicolumn{2}{c|}{\bf {it}} & \multicolumn{2}{c|}{\bf {ar}} & \multicolumn{2}{c|}{\bf {de}}  & \multicolumn{2}{c}{\bf {id}} \\
         & \bf ARC. $\uparrow$ & \bf ECE $\downarrow$ & \bf ARC. $\uparrow$ & \bf ECE $\downarrow$ & \bf ARC. $\uparrow$ & \bf ECE $\downarrow$ & \bf ARC. $\uparrow$ & \bf ECE $\downarrow$ & \bf ARC. $\uparrow$ & \bf ECE $\downarrow$ & \bf ARC. $\uparrow$ & \bf ECE $\downarrow$ \\
        \hline
        \rowcolor{platinum}
        \multicolumn{13}{c}{\textbf{{TriviaQA} on \textsl{GPT-3.5}}} \\
        \hline
       \textbf{\textit{Prob.}} & 69.58 & 30.04 & 73.21	& 46.37 & 73.08 & 28.60 & 71.51 & 46.78 & 72.48 &	33.74 & 77.37 & 50.12 \\
       \textbf{\textit{p(True)}} & 72.80 & 23.86 & 63.19 & 40.66 & 70.67 & 35.47 & 63.24 & 50.55 & 78.49 & 26.16 & 66.08 & 49.81 \\
       \textbf{\textit{Verb.}} & 71.43 & 22.18 & 72.41 & 34.80 & 72.19 & 41.54 & 76.65 & 28.68 &	68.75 & 47.14 & 69.65 & 47.14 \\
    \bottomrule
    \end{tabular}}
    \caption{Experimental results of AUROC and ECE of confidence estimations on other languages on TriviaQA for LA task on GPT-3.5.}
\label{table:other_language}
\end{table*}

\section{Uncertainty Estimations}
\label{appendix:uncer}

Both \textit{confidence} and \textit{uncertainty} estimations indicate the level of assurance of a response generated by LLMs given a query and are occasionally regarded interchangeably \cite{geng2023survey}.
Uncertainty detection is essential for hallucination mitigation on knowledge-based tasks \citep{xiong2024can,varshney2023stitch,wang2024enhancinglargelanguagemodels,vazhentsev-etal-2023-efficient,wang2024rolepromptingguideddomain,manakul-etal-2023-selfcheckgpt}.
To alleviate over-confidence and enhance the reliability of LLMs, reliable uncertainty estimation is essential to determine whether a question is known or not to the LLM \citep{geng2023survey}.
Both \textit{Uncertainty} and \textit{Confidence} estimations can indicate the reliability degree of the responses generated by LLMs, and are generally used interchangeably \cite{xiao-etal-2022-uncertainty,chen2023quantifying,geng2023survey}.
In this part, we investigate several commonly used \textit{confidence} \& \textit{uncertainty} estimation methods for generative LLMs as mentioned in Sec. \ref{sec:related}.
Specifically, we denote $\mathsf{Conf}(\boldsymbol x, \boldsymbol y)$ as the confidence score associated with the output sequence $\boldsymbol y=[y_1, y_2, \dots, y_N]$ given the input context $\boldsymbol x=[x_1, x_2, \dots, x_M]$.
We also illustrate the summarized estimation methods as well as their disadvantages in Fig. \ref{fig:conf}.

\paragraph{Likelihood-based Methods:}

Following model calibration on classification tasks \citep{guo2017calibration},
\citet{vazhentsev-etal-2023-efficient,varshney2023stitch,wang2024selfdc} intermediately quantify sentence uncertainty over token probabilities.
In traditional discriminative models, except likelihood-based methods, confidence estimations also include ensemble-based and Bayesian methods \citep{balaji2017advances,pmlr-v48-gal16,xue2022bayesian,wang2020bayesian,gal2016uncertainty,abdar2021review,9414046}, and density-based methods \citep{NEURIPS2018_abdeb6f5}.
However, this likelihood-based method requires access to token probabilities and thus being limited to white-box LLMs.
The likelihood-based confidence is estimated by calculating the joint token-level probabilities over $\boldsymbol y$ conditioned on $\boldsymbol x$.
As longer sequences are supposed to have lower joint likelihood probabilities that shrink exponentially with length, the product of conditional token probabilities of the output should be normalized by calculating the geometric mean by the sequence length \cite{murray-chiang-2018-correcting,malinin2021uncertainty}, and the confidence score can be represented as:

\begin{align}
    \mathsf{Conf}(\boldsymbol x, \boldsymbol y)=\left ( {\prod^N_ip(y_i|\boldsymbol y_{<i},\boldsymbol x)} \right )^{\frac{1}{N}}
\end{align}

\vspace{-0.5em}

Similarly, the arithmetical average of the token probabilities is adopted in \citet{varshney2023stitch}:

\vspace{-0.5em}
\begin{align}
    \mathsf{Conf}(\boldsymbol x, \boldsymbol y)=\frac{1}{N}{\sum^N_ip(y_i|\boldsymbol y_{<i},\boldsymbol x)} 
\end{align}

\vspace{-0.5em}

Furthermore, a low probability associated with even one generated token may provide more informative evidence of uncertainty \cite{varshney2023stitch}.
Hence, the minimum of token probabilities is also employed.

\begin{align}
    \mathsf{Conf}(\boldsymbol x, \boldsymbol y) = {\min \left \{ p(y_1|\boldsymbol x), \dots, p(y_N|\boldsymbol y_{<N}, \boldsymbol x) \right \}}
\end{align}
\vspace{-0.5em}

\paragraph{Prompting-based Methods:}
Recently, LLMs' remarkable instruction-following ability \citep{brown2020language} provides a view of instructing LLMs to self-estimate their confidence level to previous inputs and outputs including expressing uncertainty in words \citep{lin2022teaching,zhou-etal-2023-navigating,tian-etal-2023-just,xiong2024can}, or instructing the LLM to self-evaluate its correctness on $p(\mathrm{True})$ \citep{kadavath2022language}.
The $P(\mathrm{True})$ confidence score is implemented by simply asking the model itself if its first proposed answer $\boldsymbol y$ to the question $\boldsymbol x$ is true \cite{kadavath2022language}, and then obtaining the probability $p(\mathrm{True})$ assigned by the model, which can implicitly reflect self-reflected certainty as follows.

\vspace{-0.5em}
\begin{align}
    \mathsf{Conf}(\boldsymbol x, \boldsymbol y)=p(\mathrm{True})=p(\boldsymbol y\ \mathrm{is}\ \mathrm{True}?|\boldsymbol x)
\end{align}

Another method is to prompt LLMs to linguistically express tokens of confidence scores in verbalized numbers or words \cite{lin2022teaching,mielke-etal-2022-reducing,zhou-etal-2023-navigating,tian2023just,xiong2024can}.

The sampling-based method refers to randomly sampling multiple responses given a fixed input $\boldsymbol x$ using beam search or temperature sampling strategies \citep{manakul-etal-2023-selfcheckgpt,xiong2024can,lyu2024calibrating}.
Various aggregation methods are adopted on sampled responses to calculate the consistency level as the confidence score.
\citet{kuhn2023semantic} proposes semantic entropy to quantify uncertainty for sequences with shared meanings at the semantic level.
Moreover, some uncertainty quantification methods are used to calculate the entropy indicating the dispersion level of multiple outputs \citep{kuhn2023semantic,lin2023generating}.

\paragraph{Training-based Methods:}
For training methods, an external evaluator trained on specific datasets is introduced to output a confidence score given an input and an output.
The evaluator can be a pre-trained NLI model \citep{mielke-etal-2022-reducing}, or a value head connected to the LLM output layer \citep{lin2022teaching,kadavath2022language}, or the LLM itself \citep{han2024enhancing,xue2024ualignleveraginguncertaintyestimations}.

\begin{figure*}[t]
    \centering
    \includegraphics[width=0.95\textwidth]{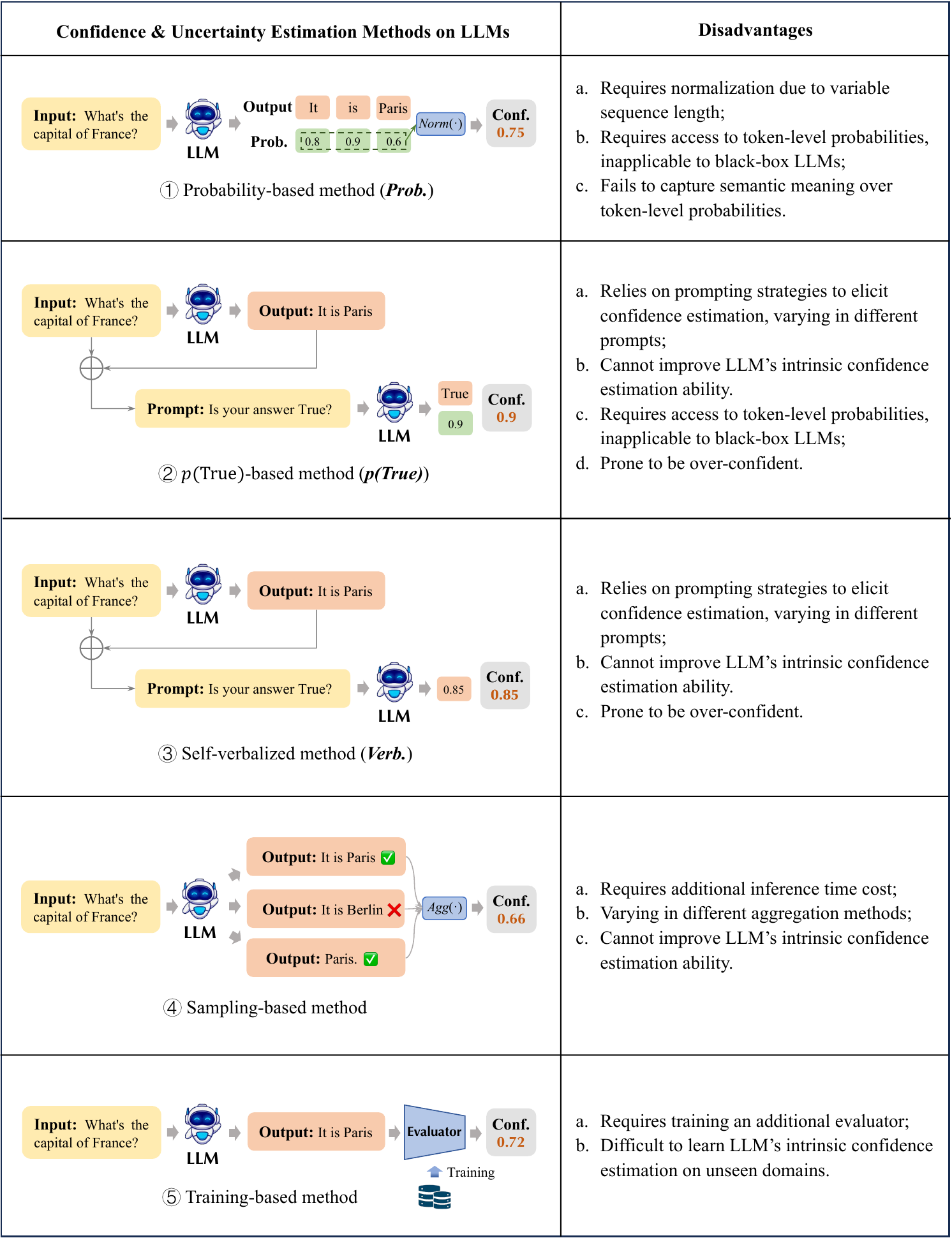}
    \caption{An illustration of several confidence estimation methods as well as their drawbacks.
    Note that sampling- and training-based methods are omitted in this work as they are cost-expensive and time-consuming for multilingual confidence estimations.
    All complete multilingual prompts used in this work are presented in Appendix \ref{appendix:prompt}.
    In addition, although \textit{confidence} and \textit{uncertainty} are always used interchangeably, the former \textit{confidence} pertains to the model's certainty regarding a specific generation, while the latter \textit{uncertainty} denotes the "dispersion" of potential predictions for a given context.
    In this work, the semantically equivalent inputs in various languages are thoroughly distinct in the token space.
    Therefore, we utilize \textit{confidence estimation} in this work, albeit specific uncertainty quantification methodologies are still applicable.}
    \label{fig:conf}
\end{figure*}

\end{document}